\documentclass[lettersize,journal,onecolumn]{IEEEtran}
\usepackage{amsmath,amssymb,amsfonts}
\usepackage{verbatim}
\usepackage{cite}
\usepackage{booktabs}
\usepackage{xcolor}
\usepackage{mathrsfs}
\usepackage{stfloats}
\usepackage{xpatch}
\usepackage{tabularx}
\usepackage{graphicx}
\usepackage{subfigure}
\usepackage{setspace}
\usepackage{algorithm}
\usepackage{algorithmicx}
\usepackage{algpseudocode}
\usepackage{multirow}
\usepackage{textcomp}
\usepackage{makecell}
\usepackage{bm}
\usepackage{pifont}
\usepackage{array} 
\graphicspath{{figures/}} 
\usepackage{lineno,hyperref}

\begin{document}

\title{A Lightweight Transfer Learning-Based State-of-Health Monitoring with Application to Lithium-ion Batteries in Autonomous Air Vehicles}

\author{Jiang~Liu, Yan~Qin,~\IEEEmembership{Member,~IEEE}, Wei~Dai,~\IEEEmembership{Senior~Member,~IEEE}, and Chau~Yuen,~\IEEEmembership{Fellow,~IEEE}
	\thanks{This work was supported by the National Natural Science Foundation of China under Grant 62373361 and Grant U24A20272, in part by the Chongqing Talent Program, and the Fundamental Research Funds for the Central Universities under Project No. 2024CDJYXTD-007 and 2024CDJZCQ-009. \textit{(Corresponding author: Yan Qin and Chau Yuen.)} }
	\thanks{Jiang Liu and Yan Qin are with the School of Automation, Chongqing
		University, Chongqing 401331, China (e-mail: liujiang2024@stu.cqu.edu.cn).}
	\thanks{Wei Dai is with the School of Information and Control Engineering, China
		University of Mining and Technology, Xuzhou 221116, China (e-mail: weidai@cumt.edu.cn).}
	\thanks{Chau Yuen is with the School of Electrical and Electronics Engineering,
		Nanyang Technological University, Singapore 639798 (e-mail: chau.yuen@ntu.edu.sg).}
}

\markboth{This paper has been accepted by IEEE Transaction on Industrial Informatics  with doi: 10.1109/TII.2025.3631012 }
{Shell \MakeLowercase{\textit{et al.}}: A Sample Article Using IEEEtran.cls for IEEE Journals}

\maketitle
\begin{abstract}
	Accurate and rapid state-of-health (SOH) monitoring plays an important role in indicating energy information for lithium-ion battery-powered portable mobile devices. To confront their variable working conditions, transfer learning (TL) emerges as a promising technique for leveraging knowledge from data-rich source working conditions, significantly reducing the training data required for SOH monitoring from target working conditions. However, traditional TL-based SOH monitoring is infeasible when applied in portable mobile devices since substantial computational resources are consumed during the TL stage and unexpectedly reduce the working endurance. To address these challenges, this paper proposes a lightweight TL-based SOH monitoring approach with constructive incremental transfer learning (CITL). First, taking advantage of the unlabeled data in the target domain, a semi-supervised TL mechanism is proposed to minimize the monitoring residual in a constructive way, through iteratively adding network nodes in the CITL. Second, the cross-domain learning ability of node parameters for CITL is comprehensively guaranteed through structural risk minimization, transfer mismatching minimization, and manifold consistency maximization. Moreover, the convergence analysis of the CITL is given, theoretically guaranteeing the efficacy of TL performance and network compactness.  Finally, the proposed approach is verified through extensive experiments with a realistic autonomous air vehicles (AAV) battery dataset collected from dozens of flight missions. Specifically, the CITL outperforms SS-TCA, MMD-LSTM-DA, DDAN, BO-CNN-TL, and AS$^3$LSTM, in SOH estimation by 83.73\%, 61.15\%, 28.24\%, 87.70\%, and 57.34\%, respectively, as evaluated using the index root mean square error.
\end{abstract}


\section{Introduction}
\label{sec:introduction}
\IEEEPARstart{T}{o} provide a safe and stable energy supply for successful ongoing tasks, lithium-ion batteries (LiBs) have been widely adopted in portable mobile devices due to the advantages of a low self-discharge rate and high energy density \cite{Qin2021Transfer}. However, LiBs' health status inevitably degrades during continuous charging and discharging cycles, caused by electrochemical side reactions over time \cite{Qin2023Atrans}. Consequently, industrial mobile devices with degraded LiBs will unexpectedly shorten the nominal service life and increase the failure risk of ongoing tasks. Therefore, accurate and rapid monitoring of the state-of-health (SOH) for LiBs has attracted extensive attention in guaranteeing safe operation for portable mobile devices \cite{shibl2023machine}.

The SOH is quantitatively evaluated by the maximum available capacity over its nominal capacity within a specific cycle \cite{xiao2025Resource}. Leveraging historical charging/discharging data, data-driven approaches for SOH monitoring hold a dominant position without the establishment of internal circuit models and electrochemical reaction mechanisms. A wide range of machine learning approaches have been reported for SOH monitoring, including back propagation network \cite{wen2022soh}, convolutional neural network (CNN) \cite{wang2024soc}, recurrent neural network \cite{10379119batteryhealth}, long short-term memory (LSTM) network \cite{zhou2023lithium}. Although the aforementioned approaches gained remarkable successes, they necessitate similar working conditions to satisfy the data distribution consistency. In practice, the changing working conditions of industrial portable mobile devices, including variable loads, various charging power, and environmental temperatures, yield inconsistent data distributions, failing the SOH monitoring model trained with a specific condition to work well under other working conditions \cite{li2020state}. As a result, the data distribution discrepancy among different working conditions challenge the performance of the existing approaches.

Transfer learning (TL) emerges as a promising paradigm to address the data discrepancy challenge between the training data in the source domain and testing data in the target domain, which usually takes advantage of the strengths of deep learning models. The basic idea of TL is to learn transferable features or network structure from the source domain. Subsequently, these features or structures are modified for the challenging tasks in the target domain. Consequently, a feasible solution is to leverage the valuable aging information in source data to assist in the training procedure of SOH monitoring model and then boost the estimator's generalization ability in the target domain with few-shot samples. For instance, Fu et al. \cite{fu2024data} pre-trained a multilayer perceptron-based source model using voltage and current data from various working conditions. The target SOH estimator is gained by fine-tuning the source model using the correlation alignment strategy, which explores the consistent characteristics across different working conditions. The strong temporal correlations within battery data motivate the wide application of LSTM for feature extraction and network construction of TL \cite{shu2021flexible}. To further enhance the transfer transparency, Ma et al. \cite{ma2023estimating} employed the maximum mean discrepancy (MMD) strategy to minimize distribution discrepancies of features and labels between source data and target data. Afterwards, a CNN is implemented for domain adaptation via updating the final fully connected layer, and the performance is verified experimentally with typical battery materials and environmental temperatures. To confront the overfitting on limited data with a fully connected layer, Lu et al. \cite{lu2024novel} used random forest to replace the fully connected layer in the pre-trained CNN source model and subsequently applied fine-tuning techniques to achieve TL-based SOH estimation across different batteries. It is worth pointing out that sufficient labeled battery data in the target domain may not be accessible in practice. On the contrary, it is affordable to collect unlabeled battery cycling data. Therefore, Li et al. \cite{li2020state} introduced a semi-supervised TL method for SOH estimation of LiBs. This approach leverages the regularization technique and MMD to realize the manifold embedding of battery data from other operating conditions, enabling high-precision estimation with limited target data. Han et al. \cite{han2022end} integrated LSTM with MMD to develop a parameter-sharing domain adaptation approach for semi-supervised SOH estimation with few-shot target cyclic data consisting of 10 randomly selected samples. Wang et al. \cite{wang2024enhanced} applied cross-entropy to establish consistent regularization between labeled and unlabeled data in the target domain for effective domain transfer.

The aforementioned battery SOH monitoring approaches achieve remarkable performance under scenarios with rich computation and storage resources, such as laboratory environments, electric vehicles, and large-scale energy storage. However, the LiB's  storable energy capacity and computation capability of portable mobile devices are quite limited, making it impossible to directly deploy these existing heavy computation resource-based TL approaches. Accordingly, portable mobile devices that prioritize low power consumption and prolonged endurance pose challenges for the aforementioned TL-based SOH monitoring approaches. Currently, no studies have investigated battery SOH monitoring in the context of resource-constrained AAVs. To this end, it is essential to develop a lightweight TL approach suitable for LiB health of state monitoring to address the following challenges:

\begin{enumerate}
	\item Limited power and computation capability carried by portable mobile devices prohibit the complex TL mechanism, calling for a lightweight SOH monitoring strategy to adapt for varying working conditions.
	\item Few labeled data in the target domain challenges the SOH monitoring performance for LiBs, requiring an efficient TL strategy to take advantage of the rich ignored unlabeled target data.
	\item Existing TL-based SOH monitoring methods lack clear interpretability in the TL process, failing to provide a solid theoretical guarantee to ensure the robust and efficient transfer capability within the SOH estimator.
\end{enumerate}

To address the above-mentioned challenges, a semi-supervised TL approach grounded in constructive incremental learning, termed constructive incremental transfer learning (CITL), is developed to meet the demand for rapid SOH monitoring for portable mobile devices with limited energy and computation resources. In particular, the proposed approach aims to build a lightweight model for accurate SOH monitoring with few-shot target data, which mainly consists of three key parts. Initially, a lightweight source estimator is established, taking the source cycling data with a regularized stochastic configuration network (RSCN). Subsequently, during the node incremental process of CITL, consistency regularization is employed to align the output mappings of RSCN and CITL for unlabeled data. The output weights of CITL are adjusted to minimize the discrepancies in stochastically mapped features between source and target domains in this incremental learning process. Consequently, degradation information is effectively extracted from the source domain, facilitating cross-domain transfer. Finally, the manifold regularization is incrementally integrated into the target estimator, which leverages the informative feature space of unlabeled data within the target domain's embedding space. The unlabeled target data is fully exploited, thereby improving the overall generalization performance of SOH estimator with limited samples from target domain. For the incremental construction process, the semi-supervised cross-domain learning mechanism is designed to constrain the stochastic generation of input parameters, enhancing the transfer efficiency of randomly mapped features. As such, CITL achieves a significant reduction in the SOH monitoring residual via iteratively increasing hidden nodes, ultimately culminating in a node-adaptive lightweight transfer model for LiBs of portable mobile devices. Overall, the main contributions of this work are summarized as follows:

\begin{enumerate}
	\item A lightweight TL strategy with end-to-end ability is proposed for SOH monitoring of LiBs, yielding a new avenue for exploring the role of constructive increment strategy for TL in portable mobile device applications.
	\item Unlabelled data in target domain are fully leveraged by exploiting the intrinsic geometry information of the entire target domain, integrating a graph-based manifold structure via the constructive incremental fashion.
	\item The semi-supervised cross-domain learning mechanism is proposed to constrain hidden parameters and control node numbers for effective TL. This offers the structural transparency and the theoretical guarantee for TL-based SOH monitoring with detailed convergence proof.
\end{enumerate}

The remainder of this paper is organized as follows. Section II illustrates the problem formulations of lightweight battery SOH monitoring with CITL. In Section III, the proposed CITL-based lightweight SOH monitoring approach of LiBs is introduced. Extensive experiments are carried out to evaluate the proposed method in Section IV. Finally, Section V draws the concluding remarks. Besides, the Appendix provides the theoretical proof of CITL with convergence analysis.
\section{Problem Formulation for Lightweight Battery SOH Monitoring with CITL}
\begin{figure}[!t]
	\centering{
		\subfigure[]
		{\includegraphics[width=0.4\columnwidth]{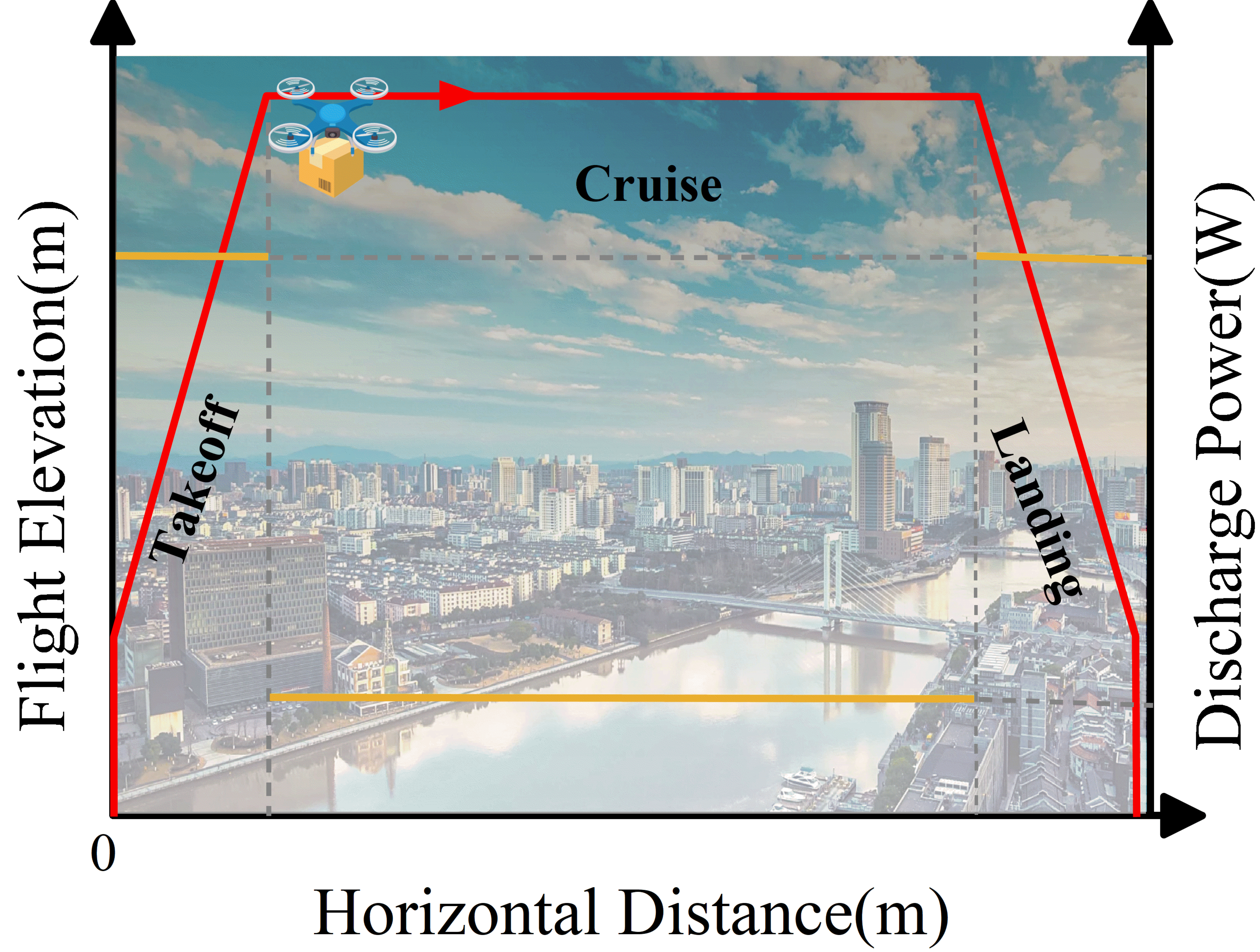}}
		\hfill
		\subfigure[]
		{\includegraphics[width=0.4\columnwidth]{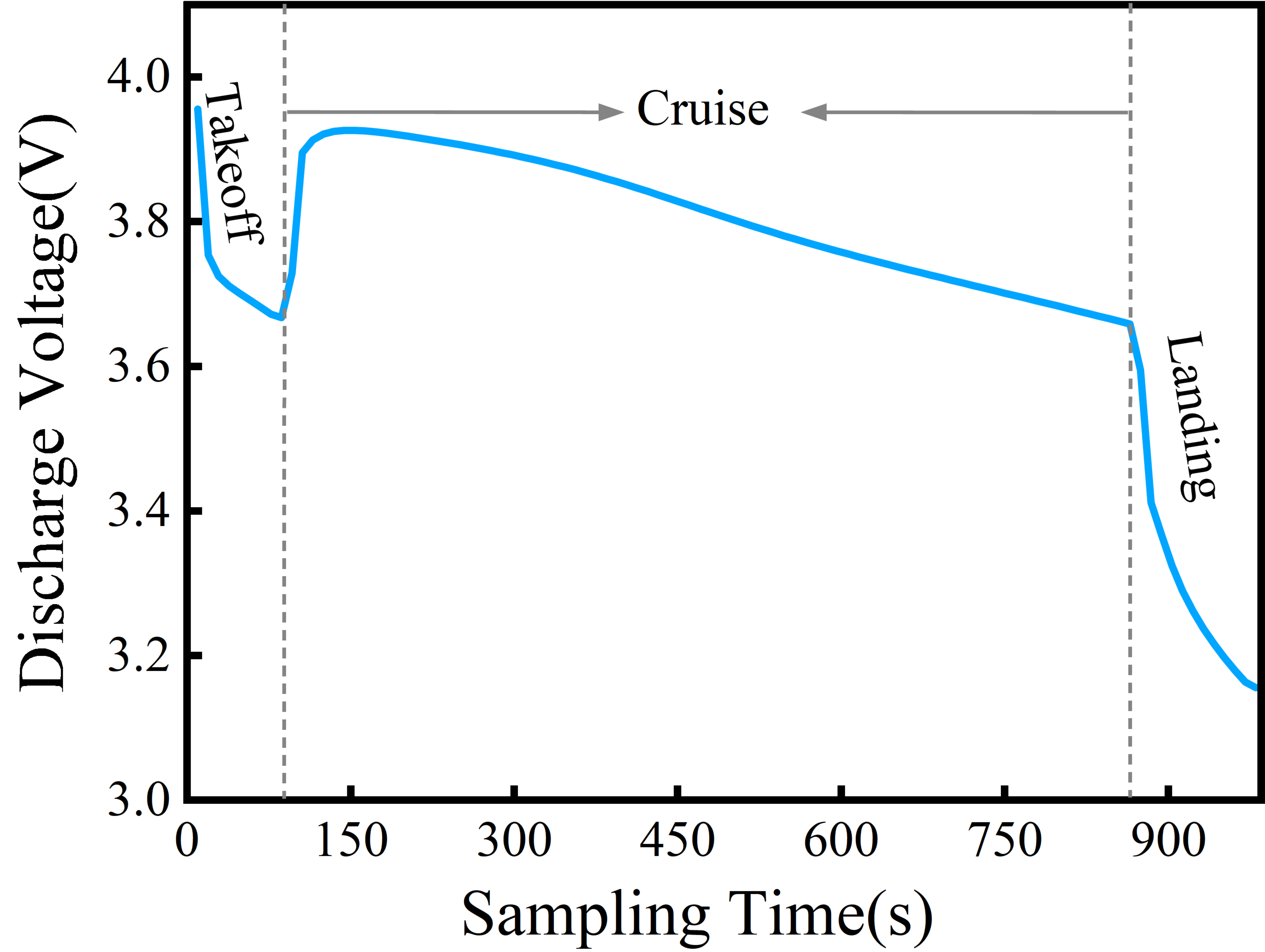}}
	}
	\caption{Illustration of flight information for an eVTOL mission cycle with (a) a three-stage flight mission and (b) corresponding discharge voltage recorded in each stage \cite{bills2023battery}.}
	\label{Fig. 1}
	\vspace{-0.4cm}
\end{figure}
\begin{figure}[!t]
	\setlength{\abovecaptionskip}{-0.1cm}
	\centering{
		\subfigure[]
		{\includegraphics[width=0.4\columnwidth]{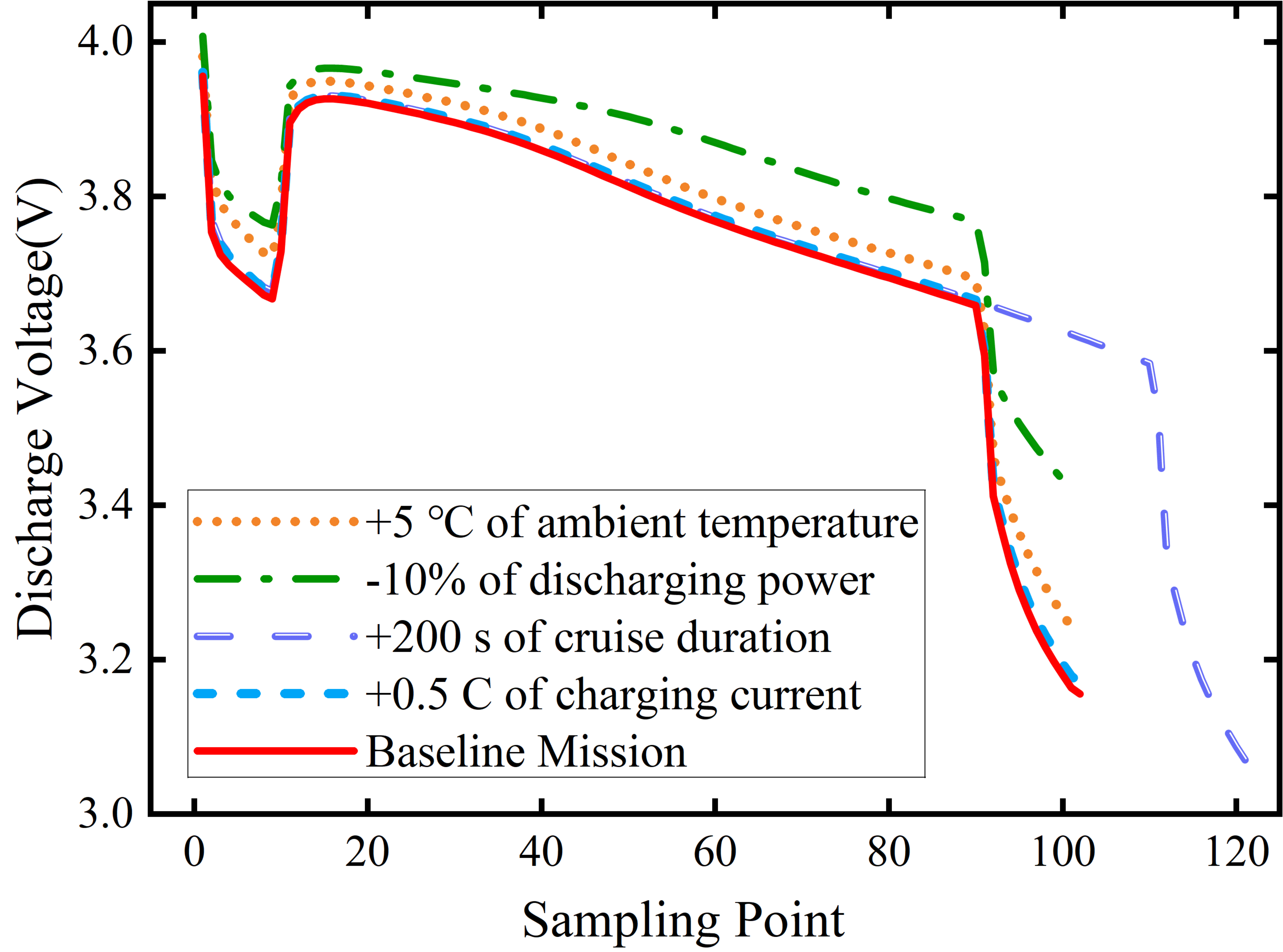}}
		\hfill
		\subfigure[]
		{\includegraphics[width=0.4\columnwidth]{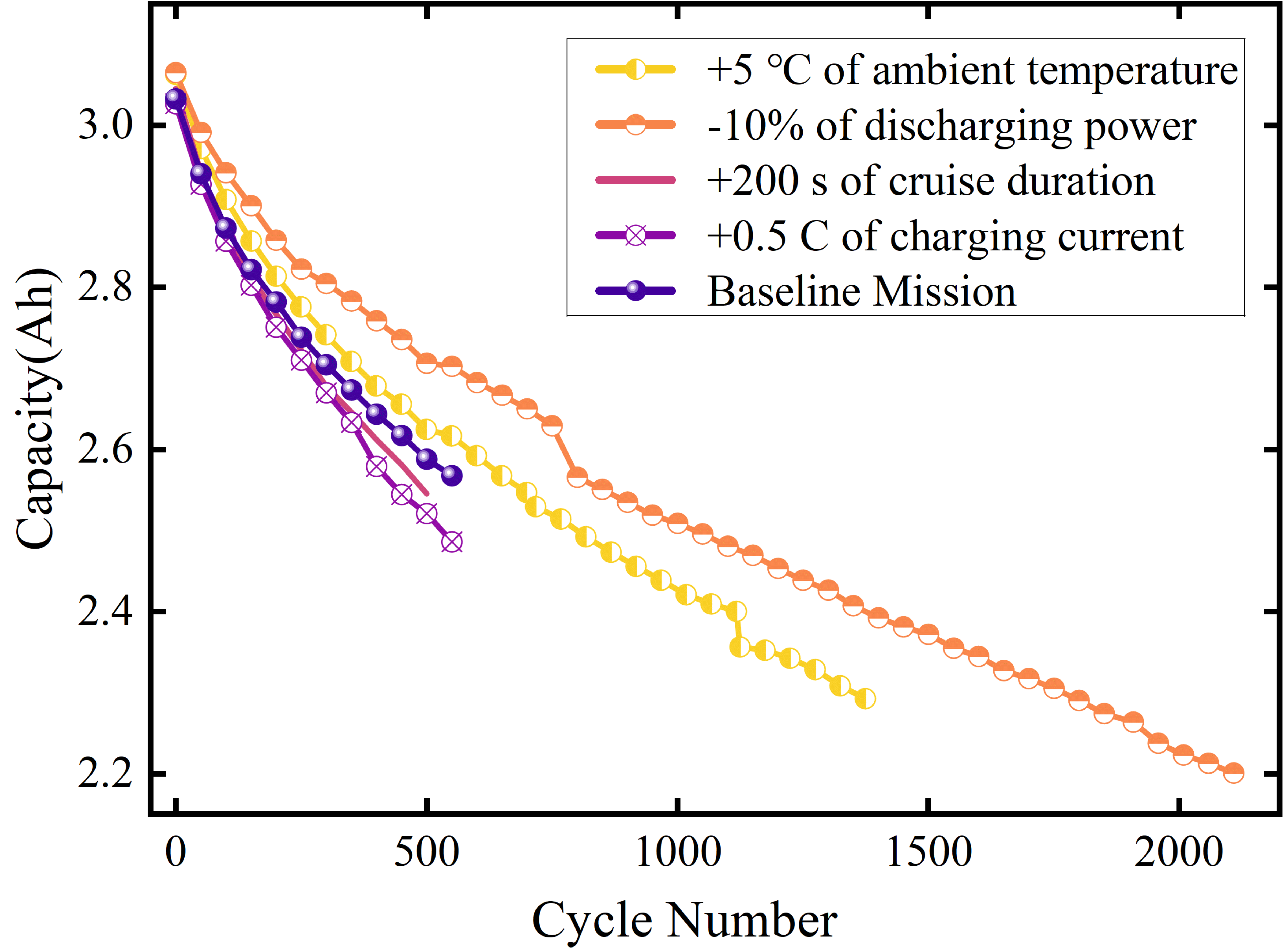}}
	}
	\caption{The measurements regarding (a) Discharge voltage curve of the first cycle and (b) Capacity degradation curve over cycles for the Sony-Murata 18650 cell of eVTOL under various working conditions.}
	\label{Fig. 2}
	\vspace{-0.6cm}
\end{figure}
This section describes the lightweight battery SOH monitoring problem formulation with CITL for achieving lightweight modeling capability. For readability, the battery data for powering an electric vertical takeoff and landing (eVTOL) aircraft, which is a typical AAV and widely utilized yet energy-limited aerial mobile device \cite{Kim2025Circular}, is briefly described in advance.

\subsubsection{Data Description}
Fig. \ref{Fig. 1}(a) briefly illustrates the typical short-range missions of an eVTOL manufactured by Vahana AAV under a specific working condition, including take-off, cruise, and landing \cite{bills2023battery}. Correspondingly, the voltage curve of the installed LiBs is demonstrated in Fig. \ref{Fig. 1}(b), which varies in response to each stage. In the take-off stage, the battery voltage drops quickly from the maximum to 3.68V with a high motor rotational speed. During the cruise stage, the discharge voltage has an obvious recovery and then reduces continuously since a low motor rotational speed drives the eVTOL flight. In the landing stage, the discharge voltage decreases quickly caused by a high motor rotational speed again.

The eVTOL is operated under a series of working conditions to perform different flight missions. In Fig. \ref{Fig. 2}(a), different working conditions of flight missions lead to significant variations in discharge voltage, including variations in time duration, amplitude, and slope, especially during the takeoff and landing stages. LiBs are continuously charged and discharged to perform repetitive missions, during which the LiBs' capacity will degrade with usage.  Fig. \ref{Fig. 2}(b) shows the LiB's capacity degradation trends under five typical working conditions, including the temperature, cruise duration, charging current, etc. Moreover, it is observed that each battery's lifespan and degradation behavior varies significantly, indicating the undeniable influence of working conditions in the discharging stage. Thus, the discharging procedure is adopted for SOH monitoring, while solely the discharging voltage is selected to reduce labor for data collection and aid in the construction of a lightweight SOH approach for AAVs.

\subsubsection{Problem Formulation}
Through introducing TL for AAV battery SOH monitoring, source domain ${{\mathcal D}_S}({\mathcal X_S},{P_S}({{\bf{X}}_S}))$ is defined as existing working conditions with a large volume of available discharging voltage data ${{\mathbf{X}}_S}$ and labeled capacities ${\mathbf{T}}_S$. $\mathcal{X}_S$ signifies the feature space of the source domain, and ${P_S}({{\mathbf{X}}_S})$ represents the marginal probability distribution about ${\mathbf{X}}_S$. Thus, a source estimator $\mathcal{F}_S( \cdot )$ with $L_S$ nodes can be modeled by minimizing the estimator error ${e_{L_S}}$ between real capacities and predictions as follows:
\begin{equation}
	{e_{{L_S}}} = \sum {{{\left[ {{{\mathbf{T}}_S} - {\mathcal{F}_S}(\mathcal{X}({{\mathbf{X}}_S}))} \right]}^2}}  \label{eq:1}
\end{equation} 

Correspondingly, target domain refers to the ongoing working condition, denoted by ${\mathcal{D}_T}({\mathcal X_T},{P_T}({{\bf{X}}_T})) = {\mathcal{D}_{Tl}} \cup {\mathcal{D}_{Tu}}$. Here, this work highlights the utilization of unlabelled discharging voltage samples ${{\mathbf{X}}_{Tu}}$ from $\mathcal{D}_{Tu}$, which is the majority of the target domain. It should be noted that discharging voltage samples ${{\mathbf{X}}_{Tl}}$ from $\mathcal{D}_{Tl}$ with labels is a small number. $\mathcal{X}_T$ signifies the feature space of target domain, and ${P_T}({{\mathbf{X}}_T})$ represents the marginal probability distribution about ${\mathbf{X}}_T$. Considering the battery's complex electrochemical mechanisms, the degradation manner in the target working condition ${\mathcal{D}_T}$ varies dynamically, challenging the development of a robust SOH monitoring model with few-shot ${{\mathbf{X}}_{Tl}}$. The difference between ${P_T}({{\mathbf{X}}_T})$ and ${P_S}({{\mathbf{X}}_S})$ prohibits the direct application of ${\mathcal{F}_{S}}( \cdot )$ for target SOH monitoring. CITL is proposed to transmit domain knowledge from source working conditions to the target domain for AAVs in a rapid and economical way. Specifically, a lightweight target estimator ${\mathcal{F}_{TL}}( \cdot )$ is anticipated to possess a broad generalization ability for ${\mathcal{D}_T}$ using an incremental construction as follows:
\begin{equation}
	{{\mathbf{T}}_T} = \sum\limits_{i = 1}^L {{\mathcal{F}_{TL,i}}\left( {{\mathcal{X}_T}({{\mathbf{X}}_T}),{\mathcal{F}_S}\left( {{\mathcal{X}_S}({{\mathbf{X}}_{Tu}})} \right)} \right)} 
\end{equation}
where $L$ denotes the node number when ${\mathcal{F}_{TL}}( \cdot )$ reaches its prescribed performance, and ${\mathcal{F}_{TL,i}}( \cdot )$ represents the output from the $i^{th}$ node within ${\mathcal{F}_{TL}}( \cdot )$. 

Consequently, our goal of accurate SOH monitoring with lightweight TL is to learn the optimal estimator ${\mathcal{F}_{TL}}( \cdot )$ by the knowledge gained from ${{\mathcal D}_S}$ and limited cycling voltages of target domain.
\section{Lightweight Battery SOH Monitoring Approach with CITL}
\begin{figure*}[t]
	\centering{
		\includegraphics[width=0.95\textwidth]{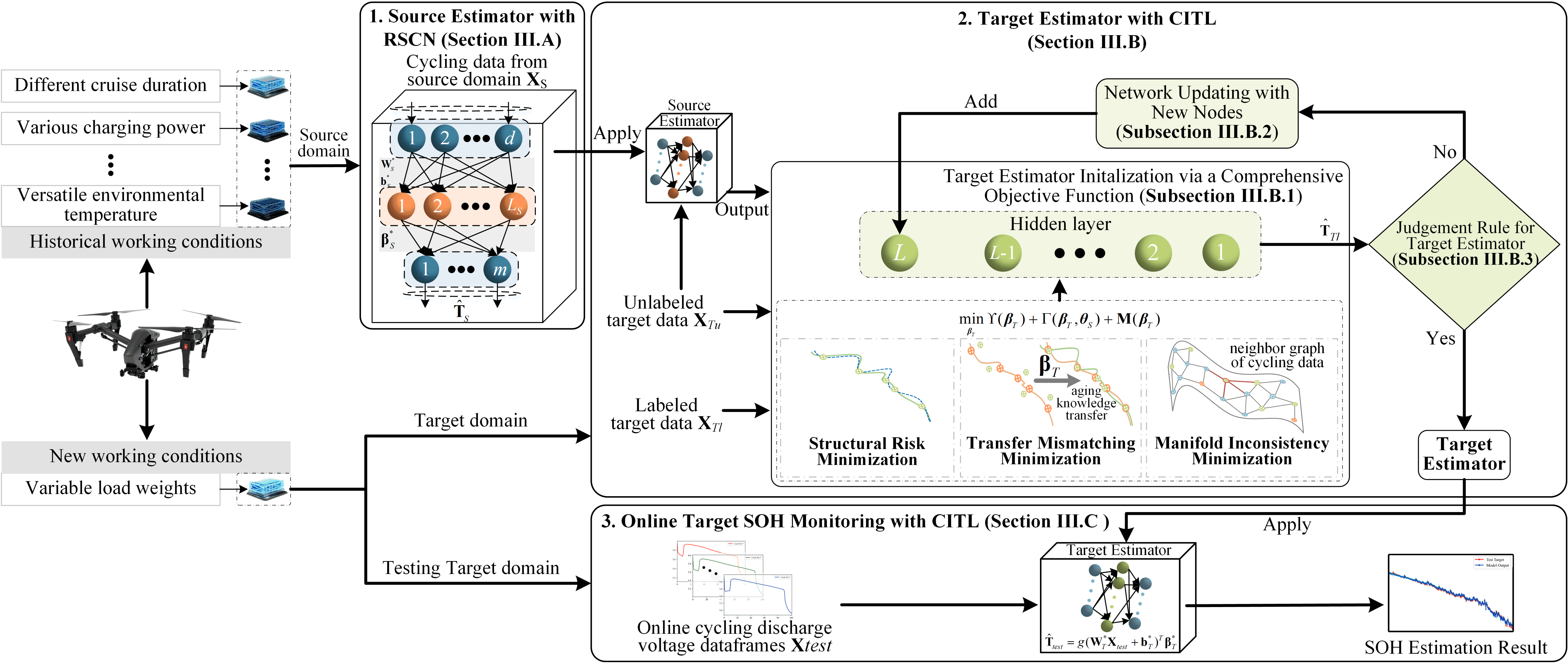}
	}
	\caption{Overall framework of the proposed CITL for lightweight TL-based SOH monitoring.}
	\label{Fig. 3}
	\vspace{-0.4cm}
\end{figure*}
In this section, the CITL is introduced for lightweight TL-based battery SOH monitoring with end-to-end ability. Its overall framework, illustrated in Fig. \ref{Fig. 3}, consists of three stages: 1) Construction of RSCN-based source estimator using cycling discharge voltage of source condition; 2) Modeling lightweight target estimator using limited cycling discharge voltage of target condition via CITL; and 3) Online SOH monitoring for target condition using well-developed CITL.
\subsection{Source Estimator with RSCN} \label{sec:sub 3.A}

RSCN \cite{wang2021underground} is equipped with a specialized supervisory mechanism that integrated with $l_2$-norm regularization of output weights compared to SCN, which effectively reduces model complexity and alleviates overfitting in few-shot scenarios \cite{Zhao2025Spatio}. Therefore, RSCN is selected for modeling source data by taking advantage of its ability regarding rapid model development and impressive generalization. This also contributes to further enhancing the lightweight characteristic of the proposed TL-based SOH monitoring method, which combines a source estimator and a target estimator.

The objective function ${\mathcal{F}_S}$ of source estimator regresses ${\mathbf{x}_S} \in {{\mathbb{R}}^d}\to \mathbf{t}_S\in{\mathbb{R}^m}$ ($m=1$) with input samples ${\mathbf{X}_{S}}=\{ {{\bf{x}}^1_{S}}, \ldots ,{{\bf{x}}^{N_S}_{S}}\} $ on corresponding labels ${\mathbf{T}_S }=\{{\mathbf{t}^1_{S}}, \ldots ,{\mathbf{t}^{N_S}_{S}}\}$. Assuming the source estimator with $L_S$-1 nodes is available, it is formulated below:
\begin{equation}
	{\mathcal{F}_{S,L_S-1}}({{\mathbf{X}}_S}) = {\sum\limits_{j = 1}^{L_S-1} {{g}\left( {{{\boldsymbol{\omega }}_{S,j}},{b_{S,j}},{{\mathbf{X}}_S}} \right)} ^T}{{\boldsymbol{\beta }}_{S,j}} \label{eq:3}
\end{equation}
where ${{\boldsymbol{\beta }}_{S,j}}$ is the output weight between the ${j}^{th}$ hidden node and output layer; ${g}( \cdot )$, $\boldsymbol{\omega}_{S,j}$, and $b_{S,j}$ are the activation function, input weight, and bias of the ${j}^{th}$ node, respectively.

\begin{figure}[t]
	\centering{
		\includegraphics[width=0.7\columnwidth]{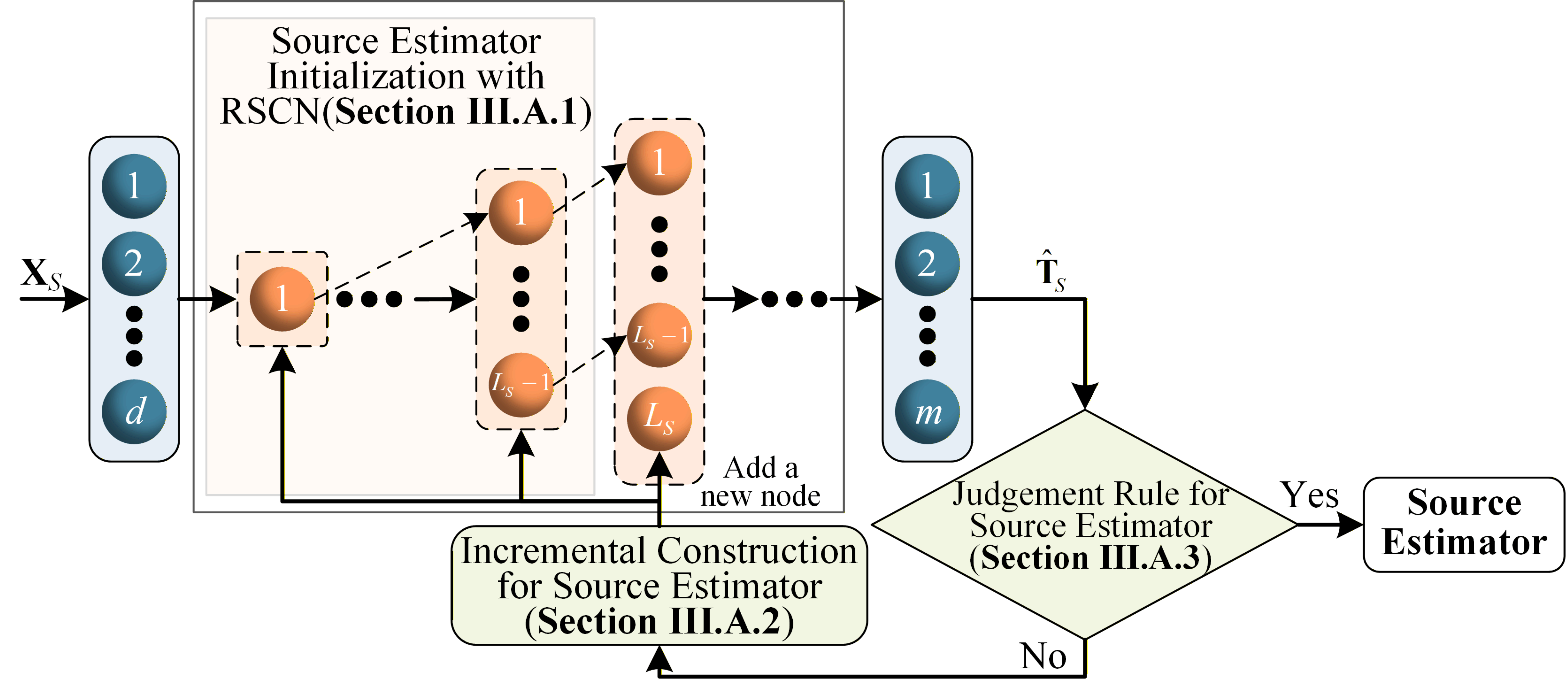}
	}
	\caption{Schematic of incremental construction for the source estimator.}
	\label{Fig. 4n}
\end{figure}

Fig. \ref{Fig. 4n} details the incremental construction modeling procedures of RSCN as follows:
\subsubsection{Source Estimator Initialization with RSCN} 
The RSCN-based source estimator is initialized with $L_S-1 \in [1, L^S_{\max}-1]$ hidden nodes:
\begin{align} 
	&\min \sum\limits_{j=1}^{L_S-1}{\left\| {{{\boldsymbol{\beta }}_{S,j}}} \right\|_2^2}  + \frac{C}{2}\sum\limits_{i=1}^{{N_S}} {\left\| {{\bf{e}}_{S,L_S-1}^i} \right\|_2^2}   \nonumber\\
	&\text{   } s.t.{\text{   }}{\bf{e}}_{S,L_S-1}^i = {\bf{t}}_S^i - {{\bf{h}}_{S,L_S-1}^i}^T{{\boldsymbol{\beta }}_{S,L_S-1}}
\end{align}
where $L^S_{\max}$ is the maximum configurable number of hidden nodes for source estimator, ${{\boldsymbol{\beta }}_{S,j}}$ has the same meaning as that in \eqref{eq:3}, ${{\bf{e}}_{S,L_S}^j}$ denotes the residual vector of SOH monitoring for $i^{th}$ sample under the current network, ${{\bf{h}}_{S,L_S}^i}$ indicates the hidden output of the $i^{th}$ node, and $C$ is the penalty coefficient.
\subsubsection{Incremental Construction for Source Estimator} \label{sec:3.A.2}
To further optimize the source estimator network, the estimation residual ${\left\| {{{\mathbf{e}}_{S,L_S-1}}} \right\|} = {\left\| {{\mathcal{F}_S} - {\mathcal{F}_{S,L_S-1}}} \right\|}$ is compared with its expected threshold $\varepsilon$. A new node ${g}({{\boldsymbol{\omega}}_{S,L_S}},{b_{S,L_S}})$ will be sequentially added to the original source estimator if $\left\| {{{\mathbf{e}}_{S,{L_S} - 1}}} \right\|$ is larger than $\varepsilon$. Correspondingly, the original source estimator is updated to ${\mathcal{F}_{S,{L_S}}} = {\mathcal{F}_{S,{L_S}}}({{\mathbf{X}}_S}) + {g}{\left( {{{\boldsymbol{\omega}}_{S,{L_S}}},{b_{S,{L_S}}},{{\mathbf{X}}_S}} \right)^T}{{\boldsymbol{\beta }}_{S,{L_S}}}$. 

The hidden output of the $L_S^{th}$ node is represented as $\mathbf{h}_{S,L_S}= {g}\left( {{\boldsymbol{\omega }}_{S,L_S}^T{\mathbf{X}_S} + {b_{S,L_S}}} \right)$. Then, $\{ {\mathbf{\tilde h}}_{S,{L_S}}^1, \ldots ,{\mathbf{\tilde h}}_{S,{L_S}}^{{T_{\max }}}\}$ is obtained by randomly generating ${{\boldsymbol{\omega}}_{L_S}}$ from ${\left[{-r,r} \right]^d}$ and ${b_{L_S}}$ from ${\left[{-r,r} \right]}$ for $T_{max}$ times. Quality factor $\wp$ is designed to search for the optimal ${{\boldsymbol{\omega}}_{L_S}}$ and ${b_{L_S}}$. For $i = 1, \ldots ,{T_{\max }}$, $\wp _i$ is denoted by:
\begin{equation}
	{\wp _i} = \frac{{{{\left\langle {{{\mathbf{e}}_{S,{L_S}}},{\mathbf{\tilde h}}_{S,{L_S}}^i} \right\rangle }^2}\left( {\left\| {{\mathbf{\tilde h}}_{S,{L_S}}^i} \right\|_2^2 + \frac{2}{C}} \right)}}{{{{\left( {\left\| {{\mathbf{\tilde h}}_{S,{L_S}}^i} \right\|_2^2 + \frac{1}{C}} \right)}^2}}} - {\mathchar'26\mkern-10mu\lambda _{{L_S}}}
\end{equation}
where ${\mathchar'26\mkern-10mu\lambda _{{L_S}}} = \left( {1 - r - {\mu _{L_S}}} \right){\left\| {{{\bf{e}}_{S,L_S}}} \right\|_2^2}$, $r \in (0,1)$, and ${\mu _{L_S}} \in (0, 1 - r)$.

The optimal hidden parameters (${{\boldsymbol{\omega}}^*_{L_S}}, {b^*_{L_S}}$) are yielded from ${\wp _i} \geqslant 0$ and ${\wp _i} \geqslant {\wp _{j \ne i}} (j=1,...,T_{max})$. Subsequently, the output weights ${\boldsymbol{\beta}^*_S}$ are globally evaluated by the regularized least squares approach \cite{wang2022multitarget}, as follows:
\begin{equation}
	\boldsymbol{\beta}_S^* = \mathop {{\arg \min} }\limits_{{\boldsymbol{\beta }}_S} \left( {\left\| {{{\boldsymbol{\beta }}_S}} \right\|_F^2 + \frac{C}{2}\left\| {{{\bf{T}}_S} - {\bf{H}}{_{S,L_S}^*}^T{{\boldsymbol{\beta }}_S}} \right\|_F^2} \right)
\end{equation}
where ${\left\|  \cdot  \right\|_F}$ represents the Frobenius norm, and ${{\bf{H}}_{S,L_S}^*} = \left[ {{{\bf{h}}_{S,1}^*}, \ldots ,{{\bf{h}}_{S,L_S}^*}} \right]^T \in {\mathbb{R}^{L_S\times N_S}}$ is hidden output of RSCN.
\subsubsection{Judgement Rule for Source Estimator} The source estimator is completed if it satisfies either of the following rules: if the node number reaches $L^S_{max}$, the construction procedure stops and outputs the final source estimator; or if $\left\| {{{\mathbf{e}}_{S,L_S}}} \right\|$ is smaller than $\varepsilon$, the source estimator is  fully established. Otherwise, the source estimator needs to be further updated according to the previous steps. 

Noteworthy, as an algorithm for globally calculating the output weights of RSCN, RSC-II \cite{wang2021underground} is introduced in Section III.A.2 for a more compact and rapid construction of the source estimator. Its predictive output can be ${{{\mathbf{\hat T}}}_S}={\bf{H}}_{S,L_S}^T{\boldsymbol{\beta }}_S^*$.
\subsection{Target Estimator with CITL}
The target estimator aims to implement high-efficiency SOH monitoring with few-shot samples from the target domain. To this end, CITL is well developed by incorporating the structural risk minimization principle and the regularization theory into a constructive incremental process. As such, the lightweight nature of the TL-based SOH target estimator is preserved, as illustrated in Fig. \ref{Fig. 3}.

\subsubsection{Target Estimator Initialization via a Comprehensive Objective Function}
Assuming the objective function ${\mathcal{F}_{TL}}$ of target domain regresses ${\mathbf{x}_T} \in {{\mathbb{R}}^d} \to \mathbf{t}_T \in {\mathbb {R}}^m (m=1)$,  the CITL with \textit{L}-1 hidden nodes is initially built as follows:
\begin{equation}
	f_{TL,L-1}({\mathbf{X}_T})= \sum\limits_{i = 1}^{L-1} {g\left( {{{\boldsymbol{\omega}}_{T,i}}{\mathbf{X}}_T^T}+{b_{T,i}} \right)^T{{\boldsymbol{\beta}}_{T,i}}}.
\end{equation}
where ${{\mathbf{X}}_T} = {{\mathbf{X}}_{Tl}} \cup {{\mathbf{X}}_{Tu}}$ is the cycling dataset comprising labeled and unlabeled samples with corresponding capacities ${{\mathbf{T}}_T} = {{\mathbf{T}}_{Tl}} \cup {{\mathbf{T}}_{Tu}} $ of target domain; ${{\boldsymbol{\omega }}_{T,i}}$ and ${b_{T,i}}$ are the input weights and bias of the $i^{th}$ node of target estimator. Its current output weight is formulated as ${{\boldsymbol{\beta }}_T} = {\left[ {{{\boldsymbol{\beta }}_{T,1}},\ldots ,{{\boldsymbol{\beta }}_{T,L-1}}} \right]^T}$ for the current CITL.

A comprehensive objective function is novelly designed with three complementarily specific goals as follows:
\begin{equation}
	\mathop {\min }\limits_{{{\boldsymbol \beta }_{T}}} {\Upsilon}({{\boldsymbol \beta }_{T}}) + {\Gamma} ({{\boldsymbol \beta }_{T}},{{\boldsymbol{\theta }}_S}) + {\rm{\bf{ M}}}({{\boldsymbol \beta }_{T}})
	\label{eq:8}
\end{equation}
where ${\boldsymbol{\theta }}_S$ refers to ${{\bf{W}}^*_S},{{\bf{b}}^*_S},{\text{and}}{\,\boldsymbol{\beta }}_S^*$ learned from the RSCN-based source estimator. 

Eq. \eqref{eq:8} includes the corresponding constraints as follows:
\begin{enumerate}
	\item Minimizing the structural risk ${\Upsilon}({{\boldsymbol \beta }_{T}})$ on a few labeled data to guarantee fundamental generalization ability and control the estimator complexity in the target domain.
	\item Minimizing the domain transfer mismatching ${\Gamma} ({{\boldsymbol \beta }_{T}},{{\boldsymbol{\theta }}_S})$ using unlabeled target data to realize aging knowledge transmission from a source domain to the target domain.
	\item Minimizing the local manifold inconsistency ${\rm{\bf{M}}}({{\boldsymbol \beta }_{T}})$ underlying the marginal distribution of different cycling samples in the target domain, thereby mining latent information from target data effectively.
\end{enumerate}

\textit{\textbf{Structural Risk Minimization:}} On the basis of minimizing the estimation error for target small samples, the $l_2$-norm regularization is introduced to control the complexity of CITL. Concretely, ${\Upsilon}$ slashes the weights of unimportant mapped features and achieves a degree of attribute selection, which reduces model complexity and avoids overfitting for the target estimator. Thus, the structural risk function is detailed below: 
\begin{equation}
	{\Upsilon}({{\boldsymbol \beta }_{T}}) =\frac{{{\mathcal{C}_T}}}{2}\sum\limits_{i = 0}^{{N_{Tl}}} {\left\| {\mathbf{e}_{Tl,L-1}^i} \right\|_{2}^2}+ \frac{1}{2}\sum\limits_{i = 1}^{L-1} {\left\| {{\boldsymbol{\beta }}_{T,i}} \right\|_2^2}  
	\label{eq:9}
\end{equation}
where $\mathcal{C}_T$ is the positive regularization parameter, $\mathbf{e}_{Tl,L-1}^i \hspace{-0.3cm}= \hspace{-0.3cm}\mathbf{t}_{Tl}^i \!\!-\!\! {\bf{H}}{{_{Tl,L-1}^i}^T}{{{\boldsymbol{\beta}}}_{T,L-1}}$ denotes the estimation residual for the $i^{th}$ target cycling sample using this CITL with \textit{L}-1 nodes, and ${\bf{H}}_{Tl,L-1}^i = {[g({\boldsymbol{\omega }}_{T,1}{{\bf{x}}_{Tl}^i}^T + {b_{T,1}}), \ldots ,g({\boldsymbol{\omega }}_{T,L}{{\bf{x}}_{Tl}^i}^T + {b_{T,L-1}})]^T}$. 

\textit{\textbf{Transfer Mismatching Minimization:}} The source estimator with sound monitoring performance has been developed using sufficient labeled data, which indicates that informative degradation knowledge is embedded within the source estimator. As such, a regressor-based lightweight knowledge transfer strategy is considered and designed in a constructive incremental fashion. Specifically, the capacity aging knowledge from the source domain is transferred to the target domain by adapting the new estimator to align with the source estimator through regularization theory. This process is mathematically formulated as follows:
\begin{align}
	\Gamma ({\boldsymbol \beta _{T}},{{\boldsymbol{\theta }}_S})= \frac{{{\mathcal{C}_{Tu}}}}{2}\sum\limits_{j = 0}^{{N_{Tu}}} {\left\| { \phi _{Tu,L-1}^i \left| {{\boldsymbol{\theta}_S}} \right. - {{\bf{H}}_{Tu}^i}^ {T} {{{\boldsymbol{\beta}}}_T}} \right\|_2^2} 
	\label{eq:10}
\end{align}
where $\mathcal{C}_{Tu}$ is the regularization parameter;  
${\bf{H}}_{Tu,L-1}^i = {[g({\boldsymbol{\omega }}^{T}_{T,1}{{\bf{x}}_{Tu}^i}\!\!+ \!{b_{T,1}}),\ldots ,g({\boldsymbol{\omega }}^{T}_{T,L-1}{{\bf{x}}_{Tu}^i}\!\! + \!{b_{T,L-1}})]^T}$ denotes the hidden output for the $i^{th}$ unlabeled sample from the target domain; $\phi _{Tu}^j\left| {{\boldsymbol{\theta} _S}} \right.={\bf{H}}{{_{S,Tu}^i}^T}{{\boldsymbol{\beta }}_S}$ indicates the hidden output in the source estimator for the $i^{th}$ unlabeled target sample, and ${\bf{H}}_{S,Tu}^i$ stands for $g({\mathbf{W}^*_S}^T{{\bf{x}}_{Tu}^i} + {{\mathbf{b}^*_S}})$. 

Note that ${{\bf{H}}_{Tu,L-1}} \in {{\mathbb R}^{(L-1) \times {N_{Tu}}}}$ represents the target domain's feature space stochastically mapped by CITL. Subsequently, ${\boldsymbol{\beta }}_{T}$ is deduced to mitigate disparities in incremental cross-domain fusion and result in leveraging the underlying information of unlabeled data. Meanwhile, CITL retains battery aging information about the labeled target domain through \eqref{eq:9}. This TL strategy enables the separate modeling of source estimator and target estimator in a distributed fashion, boosting the lightweight property of target SOH monitoring.

\textit{\textbf{Manifold Inconsistency Minimization:}} Although the feature correlation between source and target domains has been explored in \eqref{eq:10}, the feature representation of labeled and unlabeled data from such a target domain has not been thoroughly investigated. The marginal distribution between the labeled data and unlabeled data in the target domain can be further explored for better parameter optimization of CITL. That is, the unlabeled data unveils the underlying truth of the target domain, e.g. the capacity degradation trend. Thus, the manifold regularization \cite{liu2024adaptive} is incorporated as follows:
\begin{equation}
	\mathbf{M}({\boldsymbol{\beta} _{T}}) = \frac{\eta }{2}\sum\limits_{i,j = 1}^{{N_T}} {{\nu _{i,j}}\left\| {P(\mathbf{t}|{{\mathbf{x}}^i_{T}}) - P(\mathbf{t}|{{\mathbf{x}}^j_{T}})} \right\|_2^2} \label{eq:11}
\end{equation}
where $\eta$ is the positive trade-off parameter; ${\nu _{i,j}}$ denotes the pair-wise similarity weight between instance ${{\bf{x}}^i_{T}}$ and ${{\bf{x}}^j_{T}}$, and ${\nu _{i,j}}$ is $\exp( - \frac{1}{2}||{\mathbf{x}}_T^i - {\mathbf{x}}_T^j|{|^2})$ if ${{\bf{x}}^i_{T}}$ and ${{\bf{x}}^j_{T}}$ are nearest neighbors, otherwise set to zero. 

The nearest neighbor points are determined by calculating their Euclidean distance. Owing to the difficulty in computing $P(\mathbf{t}|{{\bf{x}}_{T}})$, Eq. \eqref{eq:11} is typically transformed as: 
\begin{align}
	{{\bf{M}}}({\boldsymbol{\beta}_{T}})= \frac{\eta }{2}\sum\limits_{i,j = 1}^{{N_T}} {{\nu _{i,j}}\left\| {\mathbf{t}_T^i - \mathbf{t}_T^j} \right\|_2^2} =\frac{\eta }{2}tr({{\bf{T }}^T}{\bf{LT }})
	\label{eq:13}
\end{align}
where $\mathbf{t}_T^i$ and $\mathbf{t}_T^j$ are the corresponding predicted outputs about inputting ${{\mathbf{x}}^i_{T}}$ and ${{\mathbf{x}}^j_{T}}$; ${\bf{L}} = {\bf{\Lambda }} - {\bf{V}}$ is the graph Laplacian matrix, where ${\bf{V}}=[{\nu _{ij}}]_{i = 1}^{{N_T}}{\kern 1pt} _{j = 1}^{{N_T}}$ and ${\bf{\Lambda }}$ is a diagonal matrix with its diagonal element ${\mathbf{\Lambda} _{ii}} = \sum\nolimits_j {{\nu _{i,j}}}$; and ${\bf{T}}={\bf{H}}_{T,L-1}^T{\boldsymbol{\beta }}_{T}$ indicates the output matrix of the current CITL. 

Eq. \eqref{eq:13} constructs a local nearest-neighbor graph of a cycle within the geometry space of the target domain, preserving the important properties of the original target data. As a result, the consistency between the intrinsic manifold structure of the data is maximized, enabling full use of unlabeled data.
\subsubsection{Network Updating with New Nodes} \label{sec:sub 3.B.2}
\setlength{\textfloatsep}{12pt}
\begin{algorithm}[!t]
	\caption{The procedure of CITL} 
	\footnotesize
	\begin{algorithmic}[1]
		\Require
		Limited labeled training samples of target domain $\{ ({{\bf{x}}^i_{{Tl}}}, {t^i_{{Tl}}}),{{\bf{x}}^i_{{Tl}}} \in {{\bf{X}}_{Tl}},{\bf{t}}^i_{Tl} \in {{\bf{T}}_{Tl}}\} _{i = 1}^{{N_{Tl}}}$, maximum number of hidden nodes $L_{max}$, error tolerance $\varepsilon $, maximum times of random configuration $T_{max}$, regularization parameters $\left\{ {{\mathcal{C}_T},{\mathcal{C}_{Tu}},\eta } \right\}$, source model parameters $ \left\{ {{{\boldsymbol{\beta }}_S},{{\mathbf{W}}_S},{{\mathbf{b}}_S}} \right\}$, and a set of positive scalars $r = \left[ {{\gamma _{\min }}:\Delta \gamma :{\gamma _{\max }}} \right]$.
		\Ensure
		${\boldsymbol{W}^*_T=[\boldsymbol{\omega}_{T,1}^*,...,\boldsymbol{\omega}_{T,L}^*]}$, ${{\boldsymbol{b}}_T^*}=[b^*_1,...,b^*_L]$, and ${\boldsymbol{\beta }}{_{T}^*}$.
		\State Initialize ${{\mathbf{e}}_0} = {[t_{Tl}^1, \ldots,t
			_{Tl}^{{N_{Tl}}}]^T}$, $0 < \gamma < 1$, ${\mathbf{W}}$ and ${\mathbf{\Omega }}$.
		\While {$L \leqslant {L_{max}}$ and ${\left\| {{{\mathbf{e}}_0}} \right\|_F} > \varepsilon $}
		\For{$\gamma \in r$}
		\For{$t = 1,2, \ldots ,{T_{max}}$}
		\State Randomly assign ${{\boldsymbol{\omega }}_{T,L}}$ and $b_{T,L}$ from ${\left[ { - \gamma , \gamma  } \right]^d}$ and $\left[ { - \gamma , \gamma } \right]$;
		\State Calculate ${{\mathbf{h}}_{Tl,L}}$ and ${{\mathbf{h}}_{Tu,L}}$, then obtain ${\xi _{L,q}}$ using \eqref{eq:19};
		\State Set ${\mu _L} = {{(1 - r)} \mathord{\left/{\vphantom {{(1 - r)} {(L + 1}}} 				\right.\kern-\nulldelimiterspace} {(L + 1}})$; 
		\If {$\min \left\{ {{\xi _{L,1}}, \ldots ,{\xi _{L,m}}} \right\} \geqslant 0$}
		\State Save ${{\boldsymbol{\omega }}_{T,L}}$ and $b_{T,L}$ in \textbf{W}, ${\xi_L}  = \sum\nolimits_{q = 1}^m {{\xi_{L,q}}}$ in ${\mathbf{\Omega }}$;
		\Else \State go back to \textbf{Step 4};
		\EndIf
		\EndFor
		\If{\textbf{W} is not empty}
		\State Find $\boldsymbol{\omega}_{T,L}^*$ and $b_{T,L}^*$ that maximize ${\xi_L}$ in ${\mathbf{\Omega }}$;
		\State Set ${{\mathbf{H}}_{Tl,L}} = [{{\mathbf{h}}_{Tl,1}^*, \ldots ,{\mathbf{h}}_{Tl,L}^*}]$;  
		\State Break (go to \textbf{Step 22});
		\Else \State randomly set $\tau  \in (0,1 - \gamma )$, update $\gamma  = \gamma  + \tau $;
		\EndIf
		\EndFor
		\State Calculate ${\boldsymbol{\beta }}_T^* = {\left[ {{\boldsymbol{\beta }}_{T,1}^*, \ldots ,{\boldsymbol{\beta }}_{T,L}^*} \right]^T}$ using \eqref{eq:16};
		\State Calculate ${{\mathbf{e}}_{Tl,L}} = {\mathbf{H}}_{Tl,L}^T{\boldsymbol{\beta }}_T^* - {{\mathbf{T}}_{Tl}}$;
		\State Renew ${{\mathbf{e}}_0} = {{\mathbf{e}}_{Tl,L}}$ and $L = L + 1$;
		\EndWhile
	\end{algorithmic}
\end{algorithm}
To further reduce the estimation residual $\left\| {{{\mathbf{e}}_{Tl,L-1}}} \right\|$ within the desired threshold $\varepsilon$, the CITL model is continuously updated until the number of configured hidden nodes reaches the maximum configurable number ${L_{\max }}$. Therefore, an adaptive semi-supervised cross-domain learning mechanism is constructed to appropriately assign the hidden parameters and then evaluate its output weight ${\boldsymbol{\beta }}_{T,L}$ of the newly added $L^{th}$ node. This mechanism enables the updated CITL with \textit{L} nodes to achieve higher estimation accuracy with theoretical interpretability of TL. The output weight of CITL is updated as ${{\boldsymbol{\beta }}'_{T,L}}  = {\left[ {{{\boldsymbol{\beta }}_T},{\boldsymbol \beta _{T,L}}} \right]^T}$.

To induce the input parameter configuration mode in an incremental way, the CITL network updating mechanism based on single-node evaluation output weights is initially elaborated in \textbf{Theorem 1} of Appendix I. Note that ${{\boldsymbol{\beta }}_{T,L}}$ remains fixed for additional steps. However, this leads to a slow convergence rate during the constructive incremental process. Inspired by the original RSCN, a recalculation scheme for the output weights is considered. According to \eqref{eq:8} through \eqref{eq:13}, the optimization framework $\mathcal{J}$ of CITL is formulated in the matrix form as:
\begin{align}
	\mathcal{J} =& \frac{1}{2}{\left\| {{\boldsymbol{\beta }}'_{T,L}} \right\|^2_F} + \frac{{{\mathcal{C}_T}}}{2}{\left\| {{{\mathbf{T}}_{Tl}} - {\mathbf{H}}_{Tl,L}^T{\boldsymbol{\beta }}'_{T,L}} \right\|^2_F} \nonumber\\
	&+ \!\!\frac{{{\mathcal{C}_{Tu}}}}{2}{\left\| {{{\mathbf{T}}_{Tu}} - {\mathbf{H}}_{Tu,L}^T{\boldsymbol{\beta }}'_{T,L}} \right\|^2_F} \nonumber\\
	&+ \!\! \frac{\eta }{2}tr \!\left( {{{({\mathbf{H}}_{T,L}^T{\boldsymbol{\beta }}'_{T,L})}^T}{\mathbf{L}}({\mathbf{H}}_{T,L}^T{\boldsymbol{\beta }}'_{T,L})} \right) 
\end{align}
where ${{\bf{T}}_{Tu}} = {\bf{H}}_{S,Tu}^T{{\boldsymbol{\beta }}_S}$. 

The output weights can be globally updated as follows:
\begin{align}
	{\boldsymbol{\beta}'_{T,L}}^* = &{\left[ {{\boldsymbol{\beta }}_{T,1}^*,{\boldsymbol{\beta }}_{T,2}^*, \ldots ,{\boldsymbol{\beta }}_{T,L}^*} \right]^T} = {\arg\min}_{{\boldsymbol{\beta }}'_{T,L}}\mathcal{J}.
	\label{eq:15}
\end{align}

Therefore, the optimal ${{\boldsymbol{\beta }}'_{T,L}}$ is obtained below:
\begin{align}
	{\boldsymbol{\beta}'_{T,L}}^* \!\!= &{({\bf{I}} \!+ \!{\mathcal{C}_T}{{\bf{H}}_{Tl}}{{\bf{H}}^T_{Tl}} \!+\! {\mathcal{C}_{Tu}}{{\bf{H}}_{Tu}}{\bf{H}}_{Tu}^T \!+\! \eta {{\bf{H}}_{T,L}}{\bf{LH}}_{T,L}^T)^{ - 1}}\nonumber\\
	&({\mathcal{C}_T}{{\bf{H}}_{Tl}}{{\bf{T}}_{T}} + {\mathcal{C}_{Tu}}{{\bf{H}}_{Tu}}{{\bf{T}}_{Tu}}).
	\label{eq:16}
\end{align}

Let ${\bf{e}}_{Tl,L}^* = {\mathcal{F}_{TL}} - \sum\nolimits_{j = 1}^L {{{\bf{h}}_{Tl,j}^T}{\boldsymbol{\beta }'_{T,j}}^*} $, ${\bf{e}}_{Tu,L}^* = {\mathcal{F}_{TL}} - \sum\nolimits_{j = 1}^L {{{\bf{h}}_{Tu,j}^T}{\boldsymbol{\beta }'_{T,j}}^*} $, the parameter configuration using global evaluation output weights is expounded in \textbf{Theorem 2}.

\textbf{Theorem 2}: Given these positive real numbers $({\mathcal{C}_T},{\mathcal{C}_{Tu}},\eta )$, $r \in (0,1)$ and a non-negative real number sequence $\{ {\mu _L}\} $ with ${\lim _{L \to \infty }}{\mu _L} = 0$ and ${\mu _L} \!\leqslant\! 1 - r$, we firstly denote 
\begin{equation}
	\delta _L^* = \sum\limits_{q = 1}^m (1 - r - {\mu _L}){\left\| {{\bf{e}}_{Tl,L - 1,q}^*} \right\|^2}.
\end{equation}

If hidden outputs ${{\bf{h}}_{Tl,L}}$, ${{\bf{h}}_{Tu,L}}$, and ${{\bf{h}}_{T,L}}$ are generated to satisfy the following inequality:
\begin{equation}
	A_{L,q}^* + C_{L,q}^* - D_{L,q}^* - E_{L,q}^* - F_{L,q}^* - G_{L,q}^* \geqslant b_g^2\delta _{L,q}^*
	\label{eq:18}
\end{equation}
where ${b_g} = h'_{T,L} + {\left\| {{{\mathbf{h}}_{Tl,L}}} \right\|^2}$
\begin{align}
	A_{L,q}^* =& (\frac{1}{{{\mathcal{C}_T}}} + {h'_{T,L}} + {\left\| {{{\bf{h}}_{Tl,L}}} \right\|^2} + \frac{{{\mathcal{C}_{Tu}}}}{{{\mathcal{C}_T}}}{\left\| {{{\bf{h}}_{Tu,L}}} \right\|^2})eh_{Tl,L,q}^* \nonumber\\
	C_{L,q}^* =& \frac{{2{\mathcal{C}_{Tu}}}}{{{\mathcal{C}_T}}}{{h'}_{T,L}}eh_{Tl,L,q}^*\left\langle {{\bf{e}}_{Tu,L - 1,q}^*,{{\bf{h}}_{Tu,L}}} \right\rangle \nonumber\\
	D_{L,q}^* =& {\text{ }}\frac{{2\eta }}{{{\mathcal{C}_T}}}{{h'}_{T,L}}eh_{Tl,L,q}^*\left\langle {{\bf{\zeta }}_{L - 1,q}^*,{{\bf{h}}_{T,L}}} \right\rangle \nonumber\\
	E_{L,q}^* =& \frac{{2\eta {\mathcal{C}_{Tu}}}}{{\mathcal{C}_T^2}}{{h'}_{T,L}}\left\langle {{\bf{e}}_{Tu,L - 1,q}^*,{{\bf{h}}_{Tu,L}}} \right\rangle \left\langle {{\bf{\zeta }}_{L - 1,q}^*,{{\bf{h}}_{T,L}}} \right\rangle \nonumber\\
	F_{L,q}^* =& {\text{ }}{\frac{\eta }{{{\mathcal{C}_T}}}^2}{\left\| {{{\bf{h}}_{Tl,L}}} \right\|^2}{\left\langle {{\bf{\zeta }}_{L - 1,q}^*,{{\bf{h}}_{T,L}}} \right\rangle ^2} \nonumber\\
	G_{L,q}^* = &{\frac{{{\mathcal{C}_{Tu}}}}{{{\mathcal{C}_T}}}^2}{\left\| {{{\mathbf{h}}_{Tl,L}}} \right\|^2}{\left\langle {{\mathbf{e}}_{Tu,L - 1,q}^*,{{\mathbf{h}}_{Tu,L}}} \right\rangle ^2} \nonumber
\end{align}
where $eh_{Tl,L,q}^*=\left\langle \!\!{{\mathbf{e}}^*_{Tl,L - 1,q}},{{\mathbf{h}}_{Tl,L}} \!\!\right\rangle$.

Then, the output weights of hidden nodes are evaluated by \eqref{eq:15}. Finally, $\mathop {\lim \limits}_{L \to \infty } \left\| {\mathcal{F}_{TL} - \mathcal{F}_{^{TL,L}}^*} \right\| = 0$ is obtained.

\textbf{Remark}: The output weights ${\boldsymbol{\beta}'_{T,L}}^*$ of CITL undergo complete updates after the addition of a new node to the current network by solving the global optimization of the objective function. \textbf{Theorem 2} expedites the decrease of target network residual error and accelerates the convergence speed of CITL algorithm, characterized by requiring a small number of nodes. In comparison with deep TL, the proof of this theorem deepens the theoretical interpretability of the TL process, as demonstrated in Appendix II. It should be noted that the mentioned interpretability, which highlights the clarity of the model's theoretical construction process, differs from the physical interpretability of the battery degradation mechanism \cite{wang2024physics}. The proposed semi-supervised cross-domain learning mechanism, involving \eqref{eq:15} and \eqref{eq:18}, is derived from these three optimization sub-objectives in \eqref{eq:8} that respectively establish three constraints for randomly generating input parameters. The learning mechanism gives rise to an incremental design for a shallow neural network endowed with TL capability, which promotes the structural transparency of the monitoring model. Afterwards, feature random projection is conducted on both source and target domains via hidden parameter transformation, thereby achieving an effective feature representation with an incremental form in semi-supervised TL for lightweight AAV LiB SOH monitoring.

Recall \eqref{eq:18}, ${\xi _{L}}$ is introduced to evaluate the quality of the generated $L^{th}$ node for the target estimator as follows:
\begin{align}
	{\xi _{L}} = & \frac{{A_{L}^* + C_{L}^* - D_{L}^* - E_{L}^* - F_{L}^* - G_{L}^*}}{{b_g^2}} \nonumber\\ 
	&- (1 - r - {\mu _L})\left\| {\mathbf{e}_{Tl,L - 1}^*} \right\|_2^2. 
	\label{eq:19}
\end{align}

Similar to the optimal hidden parameter search mode in Section III.A.2, the ${{\boldsymbol{\omega }}^*_{T,L}}$ and $b^*_{T,L}$ for the CITL-based target estimator can be determined according to \eqref{eq:19}. The pseudo code of CITL is summarized in \textbf{Algorithm 1}, and the proofs of the aforementioned theorems are elucidated in Appendix II.
\subsubsection{Judgement Rule for Target Estimator}
\begin{table*}[!ht] 
	\centering
	\caption{Description of AAV Mission Profiles Used for Experiments}
	\resizebox{0.9\textwidth}{!}{ \footnotesize
		\begin{tabular}{cccccccccc}
			\toprule
			\multirow{2}{*}{\textbf{Cell}} & \multirow{2}{*}{\textbf{VAH}} &\multirow{2}{*}{\textbf{Mission count}} &\multirow{2}{*}{\textbf{Working temperature}} & \multirow{2}{*}{\textbf{Cruise duration}}& \multicolumn{2}{c}{\textbf{Charge protocol}}&\multicolumn{3}{c}{\textbf{Discharge power}}\\ \cmidrule(lr){6-7}\cmidrule(lr){8-10}
			&&&&&\textbf{Constant current}&\textbf{Constant voltage}&\textbf{Takeoff} & \textbf{Crusie}& \textbf{Landing} \\
			\midrule 
			B1 & 10 &1431& 30 ℃& 800 s&1 C &4.2 V &54 W& 16 W& 54 W \\
			B2  & 11&2249 &25 ℃& 800 s&1 C &4.2 V &43.2 W&14.4 W&43.2 W \\
			B3 & 15& 554&25 ℃&1000 s&1 C &4.2 V &54 W& 16 W& 54 W \\
			B4  & 16&557 &25 ℃& 800 s&1.5 C &4.2 V &54 W& 16 W& 54 W   \\
			B5 & 17 &1002 &25 ℃& 800 s&1 C &4.2 V &54 W& 16 W& 54 W   \\
			B6 & 20 &611&25 ℃& 800 s&1.5 C &4.2 V &54 W& 16 W& 54 W  \\
			B7  & 23 & 697&25 ℃& 800 s&1 C &4.0 V&54 W& 16 W& 54 W   \\
			B8 & 24 &801 &25 ℃& 800 s&0.5 C&4.2 V &54 W& 16 W& 54 W  \\
			B9 & 25 & 554&20 ℃& 800 s&1 C &4.2 V &54 W& 16 W& 54 W   \\
			B10  & 27 &587& 25 ℃& 800 s&1 C &4.2 V &54 W& 16 W& 54 W  \\
			\bottomrule
			\multicolumn{10}{l}{VAH denotes the battery designation in the original dataset.}
	\end{tabular}}
	\label{Table 1}
\end{table*}
\begin{figure*}[!ht]
	\setlength{\abovecaptionskip}{-0.2cm}
	\centering{
		\includegraphics[width=1\textwidth]{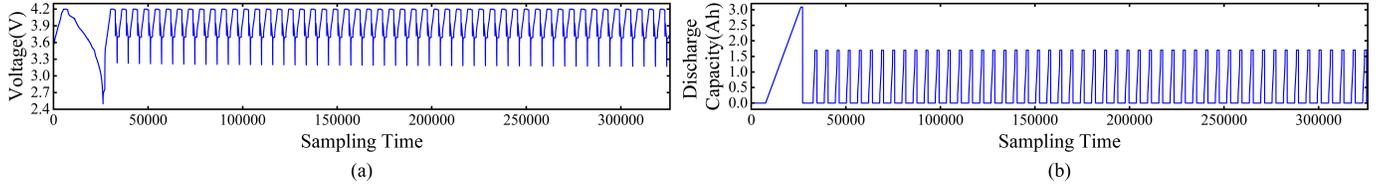}
	}
	\caption{Measurements of (a) voltage, and (b) discharge capacity for B1 cell in the first reference performance test.}
	\label{Fig. 4}
	\vspace{-0.4cm}
\end{figure*}
The target estimator completes when it satisfies the following conditions: 1) the node count reaches $L_{max}$; 2) or $\left\| {{{\mathbf{e}}_{Tl,L}}} \right\|$ is smaller than $\varepsilon$. The incremental updating procedure is stopped, and the final target estimator is produced. Otherwise, the target estimator must be further refined based on the preceding steps. 

\subsection{Online Target SOH Monitoring with CITL}
Once the CITL is well-trained offline, its input parameters (${\mathbf{W}}_T^*,{\mathbf{b}}_T^*$) and output parameters ${\boldsymbol{\beta }}_T^*$ are available. Consequently, these parameters can be used to predict the SOH online for test cycling data $\mathbf{X}_{test}$ from $\mathcal{D}_{Tu}$ of the battery target condition, demonstrated as: 
\begin{equation}
	{{\hat {\bf{T}}}_{test}} = g{({\bf{W}}{_T^*}^T{{\bf{X}}_{test}} + {\bf{b}}_T^*)^T}{\boldsymbol{\beta }}_T^*. \label{eq:37}
\end{equation}

\subsection{Computational Complexity Analysis}
The computational complexity of CITL consists of the following three parts:
\begin{enumerate}
	\item $O(L^3)$ is required to solve the linear system \eqref{eq:16} using LU decomposition.
	\item To construct the graph Laplacian matrix $\mathbf{L}$, CITL needs $O(N_TL^2)$, which is performed once.
	\item The length of $r$ and $T_{max}$ are both little predefined constant. Thus, constructing the hidden output $\mathbf H$ generally needs $O(dN_TL)$ at most.
\end{enumerate}
In summary, the computational complexity of CITL with \textit{L} hidden nodes is $O(L^3+N_TL^2+dN_TL)$. 
It is noteworthy that \textit{L} is constrained by the desired tolerance and its parameter size is equal to that of a single-layer fully connected layer in the deep network, i.e., $L(d+2)$. Therefore, the computational complexity and parameter count of CITL can be contained in a small value.

\section{Experimental Study}
\begin{table*}[!t] 
	\centering
	\caption{Main Strengths and Structure of the Employed Approaches for Comparison}
	\tabcolsep=0.5pt
	\resizebox{1.0\textwidth}{!}{ \footnotesize
		\begin{tabular}{cm{5cm}<{\centering}m{2cm}<{\centering}m{2.5cm}<{\centering}m{3cm}<{\centering}m{6.5cm}<{\centering}}
			\toprule
			\textbf{Method}&\textbf{Highlight}&\textbf{TL strategy}&\textbf{Model type}&\textbf{Model structure}& \textbf{Hyperparameter}\\
			\midrule 
			SS-TCA\cite{li2020state}&Semi-supervised transfer component analysis for feature reconstruction.& MMD-based regularization&Shallow learner&Kernel ridge regression&Tradeoff parameters ($\lambda, \mu, \gamma $) and kernel width $\Delta=1$. \\		\midrule
			MMD-LSTM-DA \cite{han2022end}&Semi-supervised domain adaptation for SOH estimation with few-shot target data.&Multi-kernel MMD&Deep learner&Two LSTM layers and one MLP layer&Epoch number, batch size, penalty factor $\alpha$, learning rate (0.001), hidden size (64), and dropout rate (0.001).
			\\ 		\midrule 
			DDAN \cite{Ye2022StateofHealthEF}&Domain adversarial TL with unsupervised feature alignment for SOH estimation&Adversarial learning&Deep learner&Three bidirectional GRU and three MLP layers&Epoch number (70), batch size (20), hidden size (64), learning rate (0.001), and tradeoff parameters ($\lambda, \beta, \nu$).
			\\ 		\midrule
			BO-CNN-TL \cite{zhang2023voltage}&Bayesian optimization for pre-training based TL with limited target data.&Fine-tuning&Deep learner&Four CNN modules and two MLP layers&Epoch number, batch size, regularization factor ($\lambda$), learning rate (0.001), and dropout rate (0.001).
			\\ 		\midrule
			AS$^{3}$LSTM \cite{JIANG2024Adaptive}&Semi-supervised self-learning TL method for SOH estimation&Teacher-student model&Deep learner&Five LSTM layers and four MLP layers&Epoch number(70), batch size (20), hidden size (64), learning rate (0.001), $L_{windows}$ and weighting factor $\alpha$\\
			\bottomrule
	\end{tabular}}
	\label{Table 2}
	\vspace{-0.2cm}
\end{table*}
\begin{table*}[!t] 
	\centering
	\caption{Comparison Results Between the Proposed Approach and Existing Methods Regarding the TL-Based SOH Estimation}
	\tabcolsep=2pt
	\resizebox{1\textwidth}{!}{ 
		\begin{tabular}{c|cccc|cccc|cccc|cccc|cccc|ccccc}
			\toprule
			\multirow{2}{*}{\textbf{TL tasks}} &\multicolumn{4}{c|}{\textbf{SS-TCA}} & \multicolumn{4}{c|}{\textbf{MMD-LSTM-DA}} &\multicolumn{4}{c|}{\textbf{DDAN}} &\multicolumn{4}{c|}{\textbf{BO-CNN-TL}}&\multicolumn{4}{c|}{\textbf{AS$^{3}$LSTM}}&\multicolumn{5}{c}{\textbf{CITL (The proposed approach)}} \\
			\cmidrule(lr){2-5}\cmidrule(lr){6-9}\cmidrule(lr){10-13}\cmidrule(lr){14-17}\cmidrule(lr){18-21}\cmidrule(lr){22-26}
			&$R^2$&$RMSE$&${T_{tr}}$&${T_{te}}$
			&$R^2$&$RMSE$&${T_{tr}}$&${T_{te}}$
			&$R^2$&$RMSE$&${T_{tr}}$&${T_{te}}$
			&$R^2$&$RMSE$&${T_{tr}}$&${T_{te}}$
			&$R^2$&$RMSE$&${T_{tr}}$&${T_{te}}$
			&$R^2$&$RMSE$&${T_{tr}}$&${T_{te}}$&{$L$}\\
			\midrule
			{B1$\to $B5}&-&2.93 &12.20&3.34 &0.73\scriptsize{$\pm$0.02}&1.48\scriptsize{$\pm$0.06}&390.75 &24.30&0.61\scriptsize{$\pm$0.46}&1.14\scriptsize{$\pm$0.56}&743.91&1317&0.76\scriptsize{$\pm$0.03}&5.03\scriptsize{$\pm$2.28}&141.84 &47.55&0.72\scriptsize{$\pm$0.33}&1.24\scriptsize{$\pm$0.51}&506.56&493.23 &\textbf{0.80\scriptsize{$\pm$0.12}}&\textbf{0.81\scriptsize{$\pm$0.25}}&\textbf{0.31}&\bf{0.10}&5.15\\
			{B5$\to $B1}&0.27&2.93 &12.14 &2.73 &0.72\scriptsize{$\pm$0.05} &1.45\scriptsize{$\pm$0.12} &172.93 &54.54&0.70\scriptsize{$\pm$0.02}&1.46\scriptsize{$\pm$0.58}&809.36 &1330&0.89\scriptsize{$\pm$0.06}&2.77\scriptsize{$\pm$0.72}&77.19&75.53&0.83\scriptsize{$\pm$0.07}&1.16\scriptsize{$\pm$0.11}&652.56 &812.11 &\textbf{0.89\scriptsize{$\pm$0.04}}&\textbf{1.13\scriptsize{$\pm$0.18}}&\bf{0.46}&\bf{0.12}&6.80 \\
			{B2$\to $B5}&- &17.73 &\textbf{16.58} &3.31&0.48\scriptsize{$\pm$0.05}&2.08\scriptsize{$\pm$0.09}&899.97 &33.42&0.62\scriptsize{$\pm$0.64}&0.72\scriptsize{$\pm$0.33}&729.87&1451&0.71\scriptsize{$\pm$0.10}&12.00\scriptsize{$\pm$3.28}&177.44 &51.07&0.68\scriptsize{$\pm$0.34}&1.83\scriptsize{$\pm$0.31}& 611.04&579.68&\textbf{0.73\scriptsize{$\pm$0.17}}&\bf{0.92\scriptsize{$\pm$0.32}}&20.18&\textbf{0.92}&101.65\\
			{B5$\to $B2}&- &6.55 &\textbf{14.69} &4.84 &\textbf{0.84\scriptsize{$\pm$0.05}}&\textbf{0.95\scriptsize{$\pm$0.15}}&238.94&63.13&0.54\scriptsize{$\pm$0.38}&1.39\scriptsize{$\pm$0.68}&989.76&1598 &0.67\scriptsize{$\pm$0.02}&5.74\scriptsize{$\pm$0.99}&56.82 &179.86&0.54\scriptsize{$\pm$0.21}&3.84\scriptsize{$\pm$0.26}&700.00 &994.65&0.67\scriptsize{$\pm$0.24} &1.40\scriptsize{$\pm$0.48}&33.78 &\textbf{1.72}&68.55 \\
			{B3$\to $B5}&-&7.43&7.14&1.25 &0.74\scriptsize{$\pm$0.03}&1.46\scriptsize{$\pm$0.07}&226.02 &37.24&0.74\scriptsize{$\pm$0.45}&0.80\scriptsize{$\pm$}0.70&936.62 &1333&0.90\scriptsize{$\pm$0.03} & 1.77\scriptsize{$\pm$0.62}&38.36&44.68&0.86\scriptsize{$\pm$0.19}&0.70\scriptsize{$\pm$0.56}&631.48&554.95&\textbf{0.92\scriptsize{$\pm$0.05}}&\textbf{0.50\scriptsize{$\pm$0.19}}&\textbf{1.93}&\textbf{0.53}&38.75\\
			{B5$\to $B3}&-&9.95 &\textbf{7.42}&1.60 &0.71\scriptsize{$\pm$0.05}&1.48\scriptsize{$\pm$0.14} &239.37&34.44&0.53\scriptsize{$\pm$0.17}&1.42\scriptsize{$\pm$0.40}&921.79 &1308&0.90\scriptsize{$\pm$0.03}&1.69\scriptsize{$\pm$0.54}&55.22& 41.50&0.73\scriptsize{$\pm$0.08}&1.44\scriptsize{$\pm$0.27}&696.24 &505.14
			&\textbf{0.91\scriptsize{$\pm$0.04}}&\textbf{0.91\scriptsize{$\pm$0.19}}&12.40&\textbf{0.82}&92.70\\
			{B4$\to $B5}&0.93&0.49&7.70&1.73&0.50\scriptsize{$\pm$0.08} &2.03\scriptsize{$\pm$1.62}&118.87 &24.26 &0.94\scriptsize{$\pm$0.05}&0.42\scriptsize{$\pm$0.20}&903.98&1260&0.89\scriptsize{$\pm$0.06} &3.42\scriptsize{$\pm$0.73}&46.30&56.81&0.76\scriptsize{$\pm$0.11}&1.38\scriptsize{$\pm$0.45}&595.08&522.94 &\textbf{0.96\scriptsize{$\pm$0.02}}&\textbf{0.35\scriptsize{$\pm$0.11}}&\textbf{0.55}&\textbf{0.19}&3.60 \\
			{B5$\to $B4}&0.88 &0.84 &9.29 &1.92 &0.82\scriptsize{$\pm$0.03}&1.19\scriptsize{$\pm$0.08}&134.44 &16.16&0.92\scriptsize{$\pm$0.09}&0.55\scriptsize{$\pm$0.39}&714.55 &1039&0.89\scriptsize{$\pm$0.02}&2.11\scriptsize{$\pm$0.37}&45.16 &40.10&0.84\scriptsize{$\pm$0.07}&1.13\scriptsize{$\pm$0.10}&671.22&470.66
			&\textbf{0.95\scriptsize{$\pm$0.05}}&\textbf{0.52\scriptsize{$\pm$0.22}}&\textbf{0.17}&\textbf{0.15}&4.65\\
			{B5$\to $B6}&0.79&1.36 & 6.38&1.63 &0.81\scriptsize{$\pm$0.02}&1.14\scriptsize{$\pm$0.05}&122.79 &17.10&0.85\scriptsize{$\pm$0.13}&0.91\scriptsize{$\pm$0.33}&864.31 &1169&0.87\scriptsize{$\pm$0.01}&4.64\scriptsize{$\pm$0.98}&45.49 & 48.56&0.77\scriptsize{$\pm$0.21}&2.23\scriptsize{$\pm$0.64}&907.73 &707.04
			&\textbf{0.97\scriptsize{$\pm$0.02}}&\textbf{0.46\scriptsize{$\pm$0.17}}&\textbf{0.59}&\textbf{0.98}&8.90 \\
			{B6$\to $B5}&0.71 &1.02 & 7.70&1.81 &0.41\scriptsize{$\pm$0.08}&2.20\scriptsize{$\pm$0.15}&120.71 &16.26&0.94\scriptsize{$\pm$0.05}&0.42\scriptsize{$\pm$0.20}&815.70 &1011&0.82\scriptsize{$\pm$0.05}&7.90\scriptsize{$\pm$2.27}&49.61 &48.94&0.65\scriptsize{$\pm$0.13}&1.18\scriptsize{$\pm$0.11}&703.83 &555.31&\textbf{0.95\scriptsize{$\pm$0.03}} &\textbf{0.40\scriptsize{$\pm$0.12}}&\textbf{0.30}&\textbf{0.20}&6.10\\
			{B5$\to $B7}&-& 7.30&8.92&2.01&0.78\scriptsize{$\pm$0.05} &1.04\scriptsize{$\pm$0.12} &116.50&15.86&0.70\scriptsize{$\pm$0.09}&0.71\scriptsize{$\pm$0.16}&864.56&1426&0.64\scriptsize{$\pm$0.03}&5.35\scriptsize{$\pm$1.41}&48.51&50.19&\textbf{0.81\scriptsize{$\pm$0.15}}&\textbf{0.69\scriptsize{$\pm$0.12}}&834.63&694.31
			&0.72\scriptsize{$\pm$0.31}&1.05\scriptsize{$\pm$0.49}&\textbf{1.78}&\textbf{0.75}&104.25\\
			{B7$\to $B5}&-& 8.34&\textbf{9.81} &2.15 &0.55\scriptsize{$\pm$0.06}&1.92\scriptsize{$\pm$0.13}& 124.20&17.38&0.77\scriptsize{$\pm$0.23}&0.91\scriptsize{$\pm$0.31}&983.32 &1223& 0.79\scriptsize{$\pm$0.07}&3.31\scriptsize{$\pm$0.79}&50.37&49.04&\textbf{0.92\scriptsize{$\pm$0.16}}&\textbf{0.44\scriptsize{$\pm$0.21}}&802.96&612.34
			&0.83\scriptsize{$\pm$0.07}&0.75\scriptsize{$\pm$0.15}&14.52 &\textbf{1.37}&199.70 \\
			{B5$\to $B8}&0.81 &1.38 &10.27 & 2.65&0.79\scriptsize{$\pm$0.03}&1.22\scriptsize{$\pm$0.10}&120.29 &21.30&0.88\scriptsize{$\pm$0.16}&0.82\scriptsize{$\pm$0.14}&932.46 &1561& 0.87\scriptsize{$\pm$}0.20&4.82\scriptsize{$\pm$1.33}&58.94& 65.05&0.81\scriptsize{$\pm$0.18}&1.87\scriptsize{$\pm$0.15}&659.78 &678.16
			&\textbf{0.98\scriptsize{$\pm$0.02}}&\textbf{0.43\scriptsize{$\pm$0.18}}&\textbf{6.42} &\textbf{1.67}&118.75 \\
			{B8$\to $B5}&0.75 &0.94 &10.23 & 2.60&0.60\scriptsize{$\pm$0.06}&1.81\scriptsize{$\pm$0.13}
			&160.16 &16.93&0.88\scriptsize{$\pm$0.11}&0.54\scriptsize{$\pm$0.31}&758.32&1266&0.80\scriptsize{$\pm$0.08}&8.43\scriptsize{$\pm$3.25}&63.48&46.52&0.68\scriptsize{$\pm$0.35}&0.72\scriptsize{$\pm$0.09}&636.45 &632.87
			&\textbf{0.98\scriptsize{$\pm$0.01}}&\textbf{0.22\scriptsize{$\pm$0.08}}&\textbf{0.24} &\textbf{0.05}&7.25 \\
			{B5$\to $B9}& 0.91&0.74&8.68 &2.05 &0.76\scriptsize{$\pm$0.06}&1.34\scriptsize{$\pm$0.16}&122.11 & 14.41&0.95\scriptsize{$\pm$0.06}&0.49\scriptsize{$\pm$0.24}&761.96&1091& 0.89\scriptsize{$\pm$0.08}&3.17\scriptsize{$\pm$0.87}& 45.47&41.63&0.74\scriptsize{$\pm$0.18}&2.32\scriptsize{$\pm$0.36}&671.16&442.63
			&\textbf{0.96\scriptsize{$\pm$0.03}}&\textbf{0.46\scriptsize{$\pm$0.18}}&\textbf{0.30} &\textbf{0.17}&7.35 \\
			{B9$\to $B5}& 0.89&0.63 & 14.84& 4.83&0.44\scriptsize{$\pm$}0.08 &2.15\scriptsize{$\pm$0.15}&100.88 &15.68&0.76\scriptsize{$\pm$0.15}&0.68\scriptsize{$\pm$0.21}&878.38 &1202&0.85\scriptsize{$\pm$0.04}&5.13\scriptsize{$\pm$2.05}&52.54 &48.02&0.88\scriptsize{$\pm$0.09}&0.53\scriptsize{$\pm$0.12}&569.83 &549.31&\textbf{0.94\scriptsize{$\pm$0.04}}&\textbf{0.43\scriptsize{$\pm$0.15}} &\textbf{0.90}&\textbf{0.40}&8.25 \\
			{B5$\to $B10}&0.80 & 1.32&9.29 &1.90 & 0.78\scriptsize{$\pm$0.04}&1.22\scriptsize{$\pm$0.10}&152.76 &25.45&0.50\scriptsize{$\pm$0.48}&1.27\scriptsize{$\pm$0.27}& 793.74 &1141&0.86\scriptsize{$\pm$0.01} &5.32\scriptsize{$\pm$0.83}&52.52 &49.74&0.56\scriptsize{$\pm$0.36}&1.70\scriptsize{$\pm$0.26}&671.16 &516.17
			&\textbf{0.99\scriptsize{$\pm$0.01}}&\textbf{0.33\scriptsize{$\pm$0.10}}& \textbf{0.48}&\textbf{0.15}&7.90\\
			{B10$\to $B5}&0.73& 0.97& 8.94& 2.22&0.42\scriptsize{$\pm$0.11}&2.17\scriptsize{$\pm$0.21}&140.59 &17.68&0.87\scriptsize{$\pm$0.15}&0.53\scriptsize{$\pm$0.26}&701.96&1286&0.72\scriptsize{$\pm$0.04}&7.87\scriptsize{$\pm$1.91}&49.26 &41.15&0.57\scriptsize{$\pm$0.35}&1.68\scriptsize{$\pm$0.48}&937.23 &859.64
			&\textbf{0.95\scriptsize{$\pm$0.08}}&\textbf{0.36\scriptsize{$\pm$0.21}}& \textbf{0.23}&\textbf{0.15}&5.70 \\
			{B9$\to $B10}&0.78 &1.39 & 9.68& 2.16&0.68\scriptsize{$\pm$0.06}&1.46\scriptsize{$\pm$0.15} &128.35&20.76&0.76\scriptsize{$\pm$0.17}&1.02\scriptsize{$\pm$0.53}&721.25&1299&0.86\scriptsize{$\pm$}0.06&4.61\scriptsize{$\pm$2.74}&26.62&42.28&0.83\scriptsize{$\pm$0.11}&1.47\scriptsize{$\pm$0.20}&850.90&767.85
			&\textbf{0.98\scriptsize{$\pm$0.02}}&\textbf{0.40\scriptsize{$\pm$0.12}}& \textbf{0.29}&\textbf{0.70}&6.35\\
			{B10$\to $B9}&0.89 & 0.83& 8.46& 2.22&0.65\scriptsize{$\pm$0.07}&1.61\scriptsize{$\pm$0.17}& 141.29&17.69&0.84\scriptsize{$\pm$0.16}&0.76\scriptsize{$\pm$0.48}&694.45&1216&0.88\scriptsize{$\pm$0.02}&4.05\scriptsize{$\pm$1.23}&29.96&39.51&0.83\scriptsize{$\pm$0.05}&1.21\scriptsize{$\pm$0.18}& 905.53 &599.31&\textbf{0.96\scriptsize{$\pm$0.05}}& \textbf{0.43\scriptsize{$\pm$0.22}}&\textbf{0.22}&\textbf{0.25}&5.40 \\
			\textbf{Average}&0.47&3.75&10.03&2.45&0.66\scriptsize{$\pm$0.05}&1.57\scriptsize{$\pm$0.20}&159.73&24.36&0.80\scriptsize{$\pm$0.21}&0.85\scriptsize{$\pm$}0.36&903.68&1276&0.82\scriptsize{$\pm$0.04}&4.96\scriptsize{$\pm$}1.46&60.57&55.39&0.75\scriptsize{$\pm$0.19}&1.43\scriptsize{$\pm$0.27}&648.66&627.27
			&\textbf{0.90\scriptsize{$\pm$0.07}}&\textbf{0.61\scriptsize{$\pm$0.21}}&\textbf{4.80}&\textbf{0.57}&40.39
			\\
			\bottomrule
			\multicolumn{18}{l}{${T_{tr}}$, $T_{te}$, and $L$ are the training time (in seconds), testing time (in milliseconds), and number of used hidden nodes, respectively;}\\
			\multicolumn{18}{l}{$RMSE$ denotes the root mean square error (\%); $R^2$ is the coefficient of determination; - indicates a failure in SOH prediction with ${R^2} < 0$. }
	\end{tabular}}
	\label{Table 3}
	\vspace{-0.4cm}
\end{table*}
A series of comparisons are conducted in this section to verify the efficacy of the proposed SOH monitoring approach, specifically regarding the cross-domain learning with limited labels in target domain. The battery dataset \cite{bills2023battery}, collected under real-world cruising situations such as various charging protocols, ambient temperatures, cruise duration, and flight power, is utilized. Table \ref{Table 1} shows the 10 mission profiles with specific conditions, wherein B5 and B10 cells serve as the baseline mission. A reference performance test (RPT), involving a full charge and discharge cycle, was undertaken at the start of each aging test campaign and after each subsequent set of general 50 mission cycles. Taking B1 cell as an example, its voltage and discharge capacity measurements in the first RPT are illustrated in Fig. \ref{Fig. 4}. To obtain a substantial amount of experimental data, the cycling data after each PRT is used, and the battery capacities are calculated by subtracting the discharge capacity of each cycle from the capacity obtained from the first PRT in this paper.
\subsection{Experimental Setup}
The first 20 cycles of the target domain battery are employed as the labeled training samples, and the remaining cyclic samples are unlabeled. The first 20 unlabeled samples are used for semi-supervised training, and the remaining for testing. The testing samples encompass all cycles of the target domain. A TL task is regarded as $\mathcal{A}\to \mathcal{B}$, where $\mathcal{A}$ denotes the source domain and $\mathcal{B}$ denotes the target domain. Thus, the RSCN-based source estimator is developed using discharging voltage data from $\mathcal{A}$, while the CITL-based target estimator is developed using limited discharging voltage data from $\mathcal{B}$. The shared hyperparameters of RSCN and CITL are set as follows:  $L_{max}=200$, $T_{max}=50$, $\varepsilon=0.01 $, and $r=[0.5, 1, 5, 10, 50, 100, 200]$. The remaining parameters of CITL and those of subsequent comparison methods are configured by Bayesian optimization \cite{zhang2023voltage} with 100 trials. Experimental results, such as the mean and standard deviation, are obtained by conducting the same 20 random trials for fair comparison.

All experiments, except for those in the last subsection, are conducted on a Lenovo GeekPro-171RR computer equipped with Windows 11 system and an Intel Core i7-14700F CPU (2.10 GHz, 32 GB of RAM). Experimental models are implemented in Python 3.8 and run on the CPU for simulating the resource constraints of AAVs in real-world scenarios.
\subsection{TL Performance Evaluation in Various Conditions} \label{sec:4.1}
This subsection will evaluate and compare the TL performance of the proposed approach against existing popular TL-based SOH estimation methods under diverse conditions. Table \ref{Table 2} summarizes the basic information of the employed comparison approaches, including the TL strategy, model types, model structure, etc. The comprehensive experimental results are recorded in Table \ref{Table 3}, which also indicates the number of hidden nodes utilized by CITL. The standard deviations of $R^2$ and $RMSE$ for SS-TCA with analytical solution are zero and omitted from the table. Meanwhile, $R^2$ is used to assess the overall goodness-of-fit of the model. Specifically, $R^2<0$ implies that the model performs worse than a naive predictor that simply utilizes the mean SOH of target data as the prediction, resulting in meaningless predictions. As a whole, CITL demonstrates superior performance in TL-based SOH estimation, surpassing all other methods in 17 out of 20 TL tasks in terms of both estimation accuracy and prediction speed. For the average results across all TL tasks, CITL obtains the most accurate estimation with an $RMSE$ of 0.61\%, yielding improvements of 83.73\%, 61.15\%, 28.24\%, 57.34\%, 87.70\%, and 57.34\% over SS-TCA, MMD-LSTM-DA, DDAN, BO-CNN-TL, and AS$^3$LSTM, respectively.

Although SS-TCA is a shallow learner model with adequate prediction speed, it cannot accurately track the capacity degradation trend of the target battery with poor estimation accuracy with $R^2<0$ in most cases, such as B1$\to$B5, B2$\to$B5, B3$\to$B5, B5$\to$B3, and B7$\to$B5. It is worth note that SS-TCA necessitates the selection of a regressor for final SOH estimation following feature reconstruction, which greatly affects its estimation performance. MMD-LTSM-TL, DDAN, BO-CNN-TL, and AS$^3$LSTM, based on deep learning, consume substantial computing resources for parameter tuning, resulting in prolonged training and testing times for SOH estimation. The ${T_{tr}}$ of MMD-LSTM-DA, DDAN, and AS$^3$LSTM exceeds 100 seconds per TL task. Particularly, the $T_{te}$ of AS$^3$LSTM, constructed based on multi-layer networks, exceeds 1 second per task. However, they struggle to achieve satisfactory accuracy and prediction performance with 20 labeled target data. This is because the network structures of deep TL-based SOH estimation methods are predefined and remain fixed in each trial, where both the number of nodes and layers introduce significant uncertainty in estimation accuracy and scale up network parameters. In contrast, CITL requires less than 1 second for training in most tasks and approximately 1 millisecond for predicting per task. Meanwhile, the proposed CITL achieves satisfactory results using fewer than 10 hidden nodes in most TL tasks due to its self-controlled constructive incremental modeling.

Noteworthily, compared with the feature reconstruction method SS-TCA, CITL is not entangled with regressor selection by integrating cross-domain learning and lightweight regressor into a complementary framework. This contributes to the convenience of its application in AAVs. For LSTM-based or CNN-based TL methods with end-to-end capability, CITL also possesses the end-to-end functionality within a random weight neural network, yet a unique lightweight TL strategy for SOH estimation. Therefore, the CITL is well-suited for SOH estimation of AAV batteries with few-shot data, such as with only 20 labeled samples.
\subsection{Node Increment Analysis of CITL}
\begin{figure*}[!ht]
	\centering{
		\subfigure[]
		{\includegraphics[width=0.37\columnwidth]{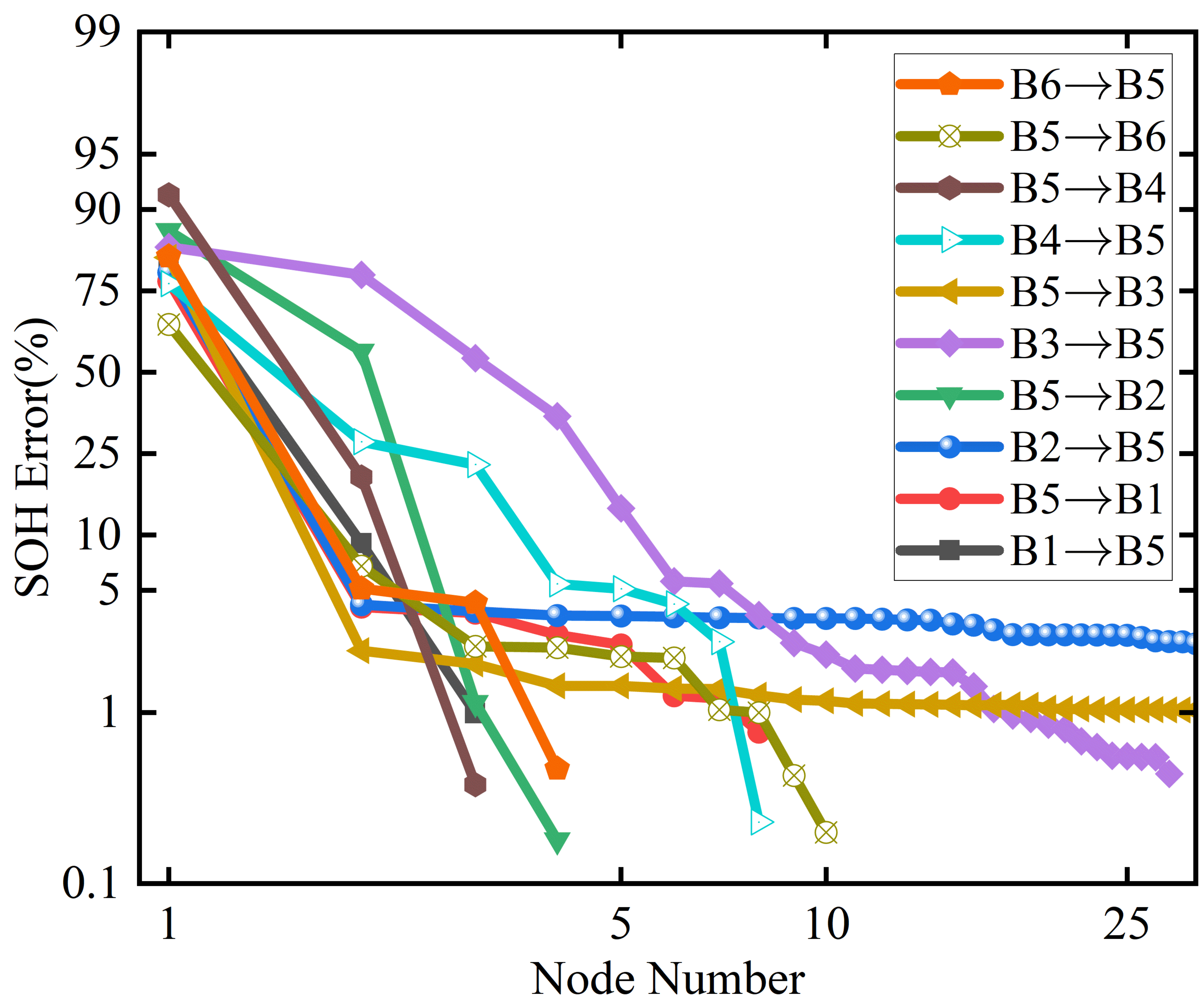}}
		\hfill
		\subfigure[]
		{\includegraphics[width=0.55\columnwidth]{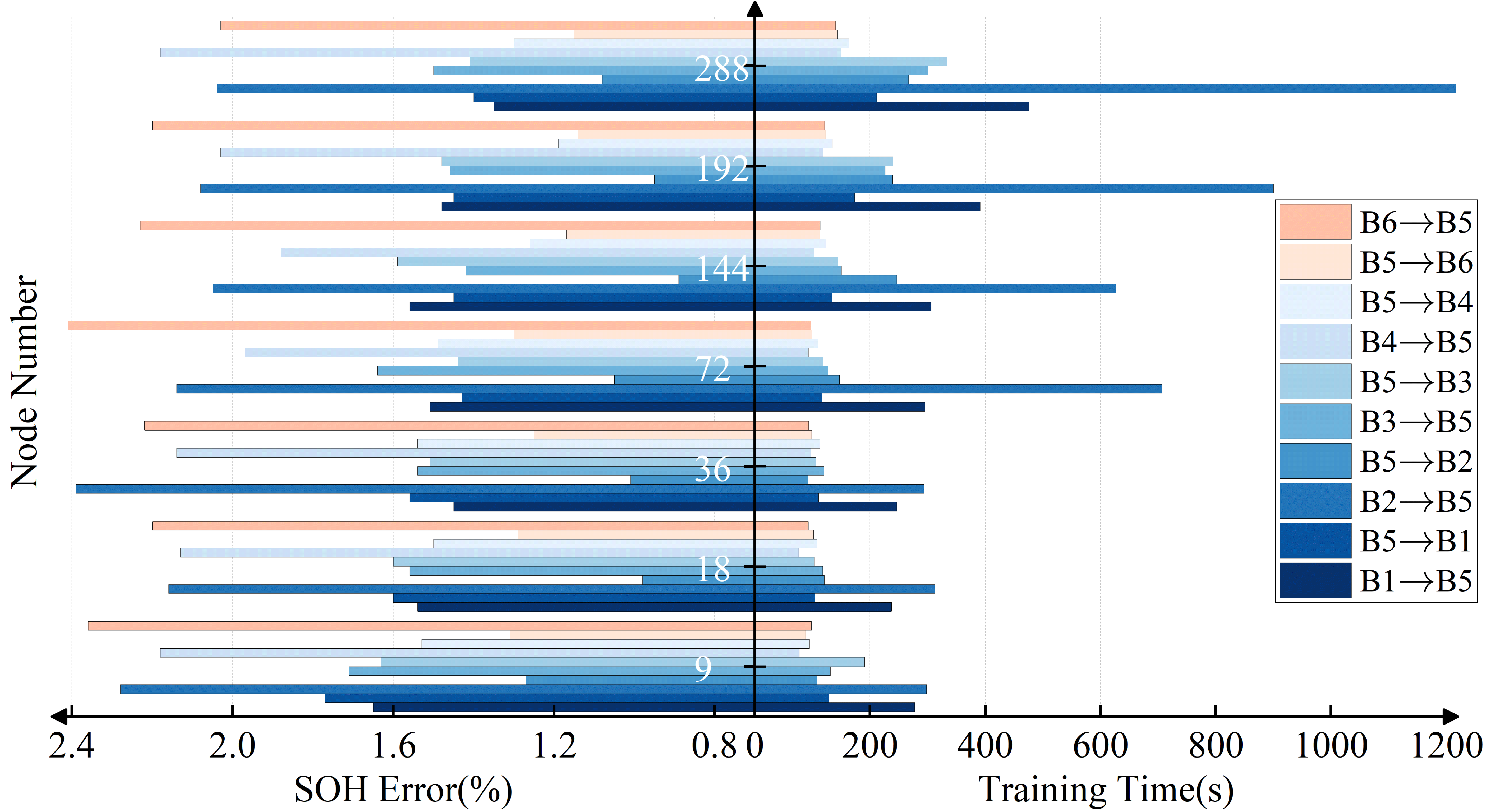}}
	}
	\caption{Impact of node number on the (a) SOH estimate error for CITL, and (b) SOH estimate error and training time for MMD-LSTM-DA.}
	\label{Fig. 5}
	\vspace{-0.2cm}
\end{figure*}
\begin{figure}[!t]
	\centering{
		\includegraphics[width=0.6\columnwidth]{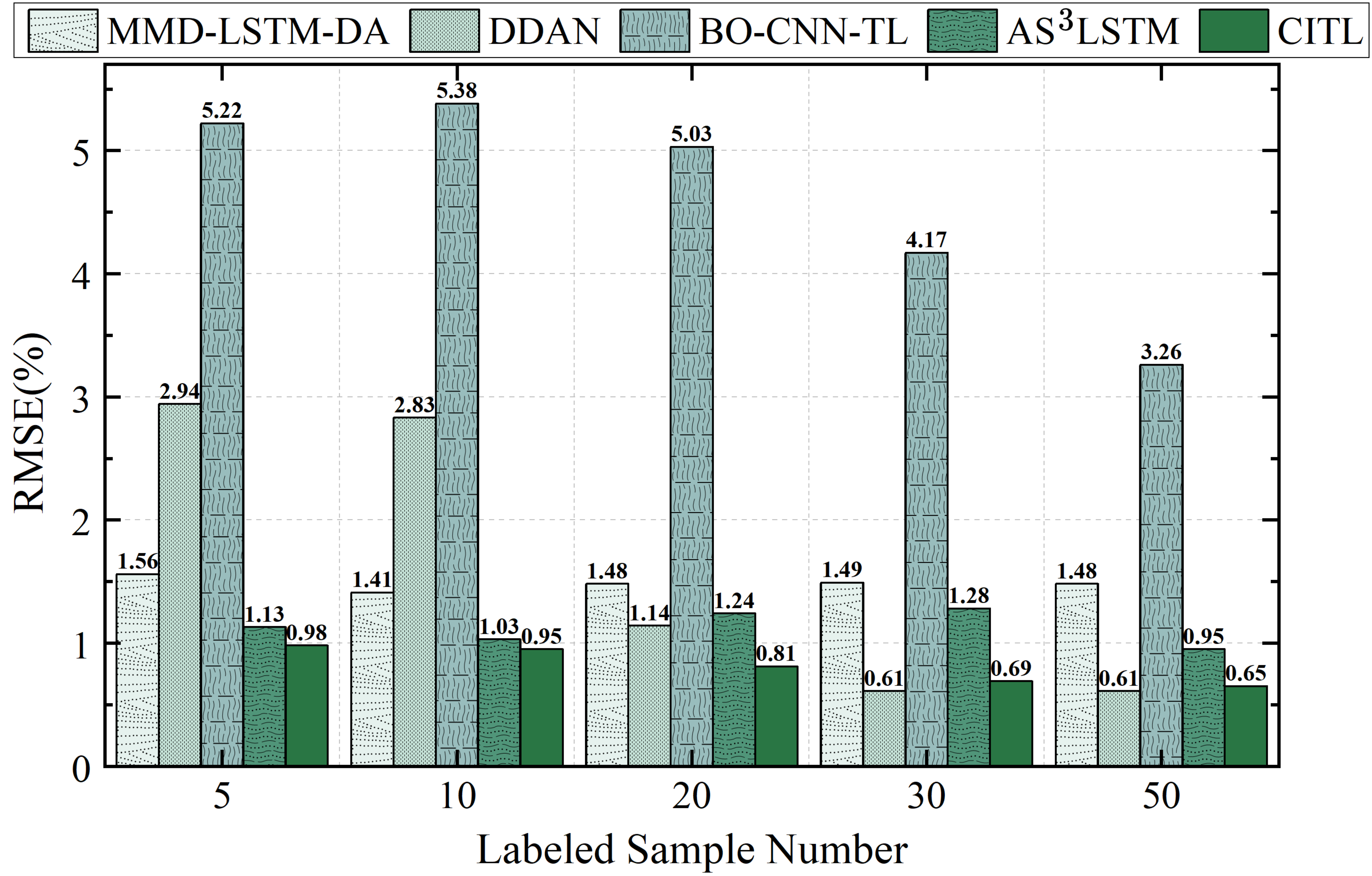}
	}
	\caption{Estimation error comparison between CITL and its counterparts with several different labeled samples.}
	\label{Fig. 88}
\end{figure}
\begin{figure*}[!ht]
	\centering{
		\subfigure[]
		{\includegraphics[width=0.3\textwidth]{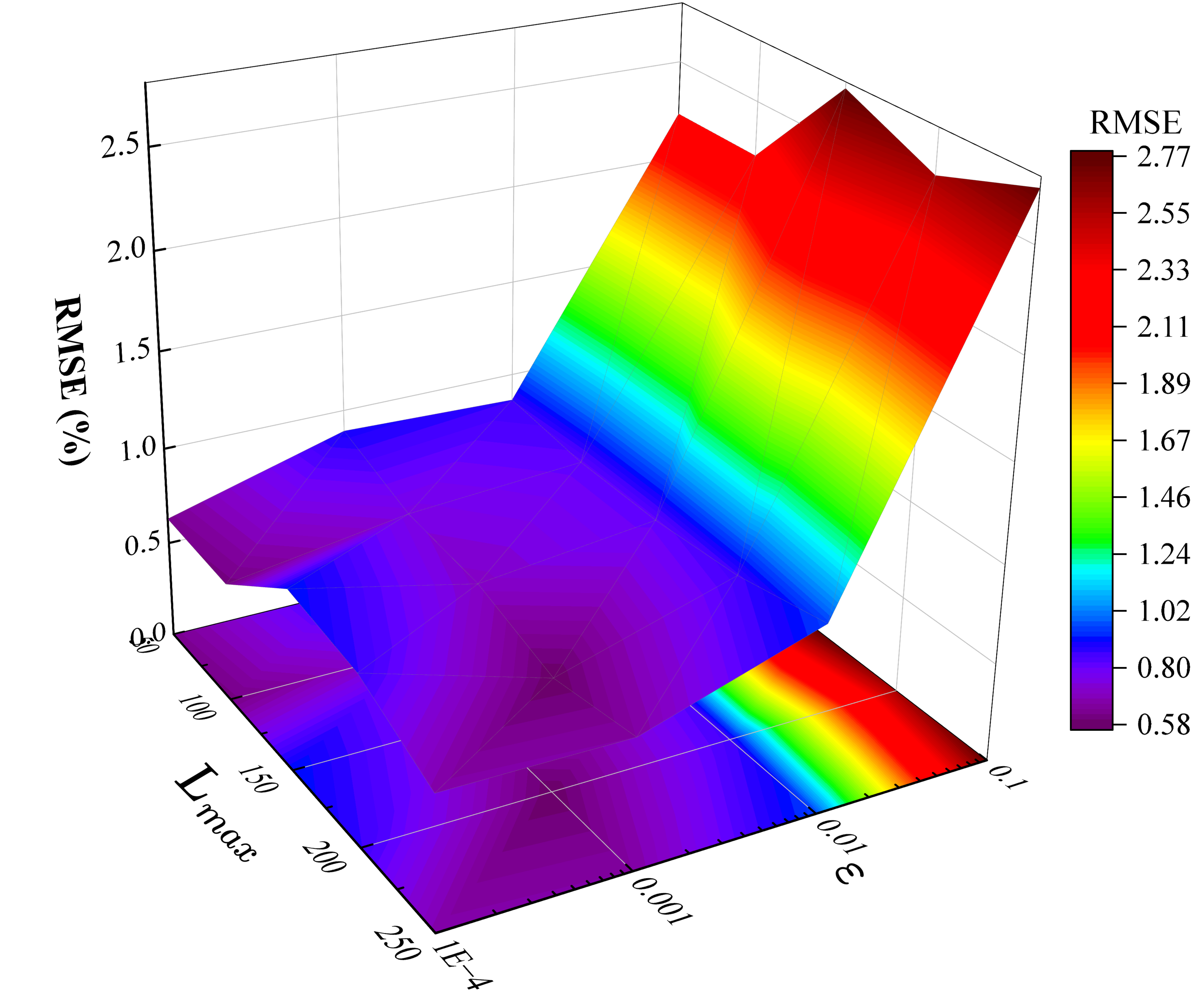}} \hfill
		\subfigure[]
		{\includegraphics[width=0.3\textwidth]{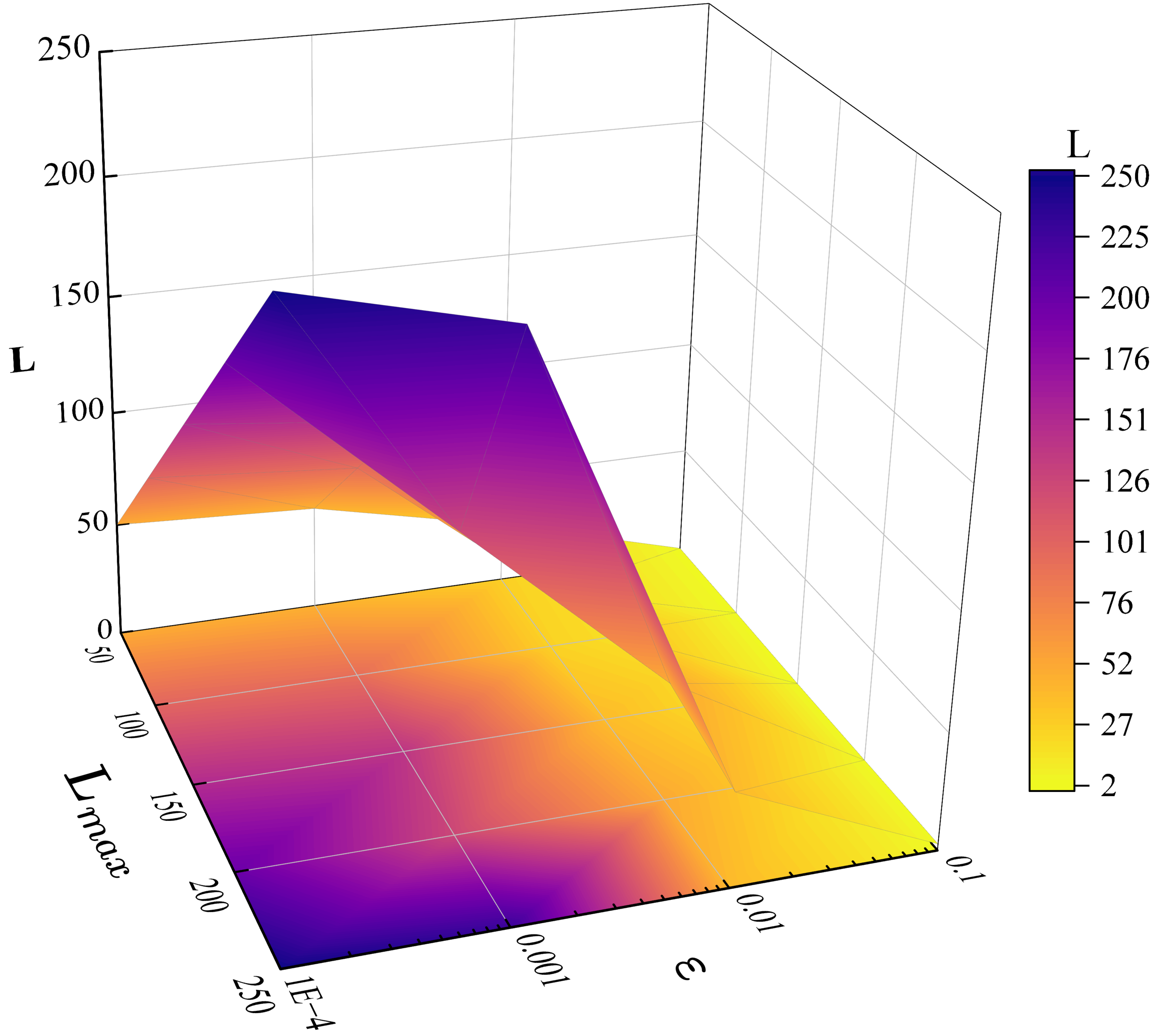}} \hfill
		\subfigure[]		
		{\includegraphics[width=0.33\textwidth]{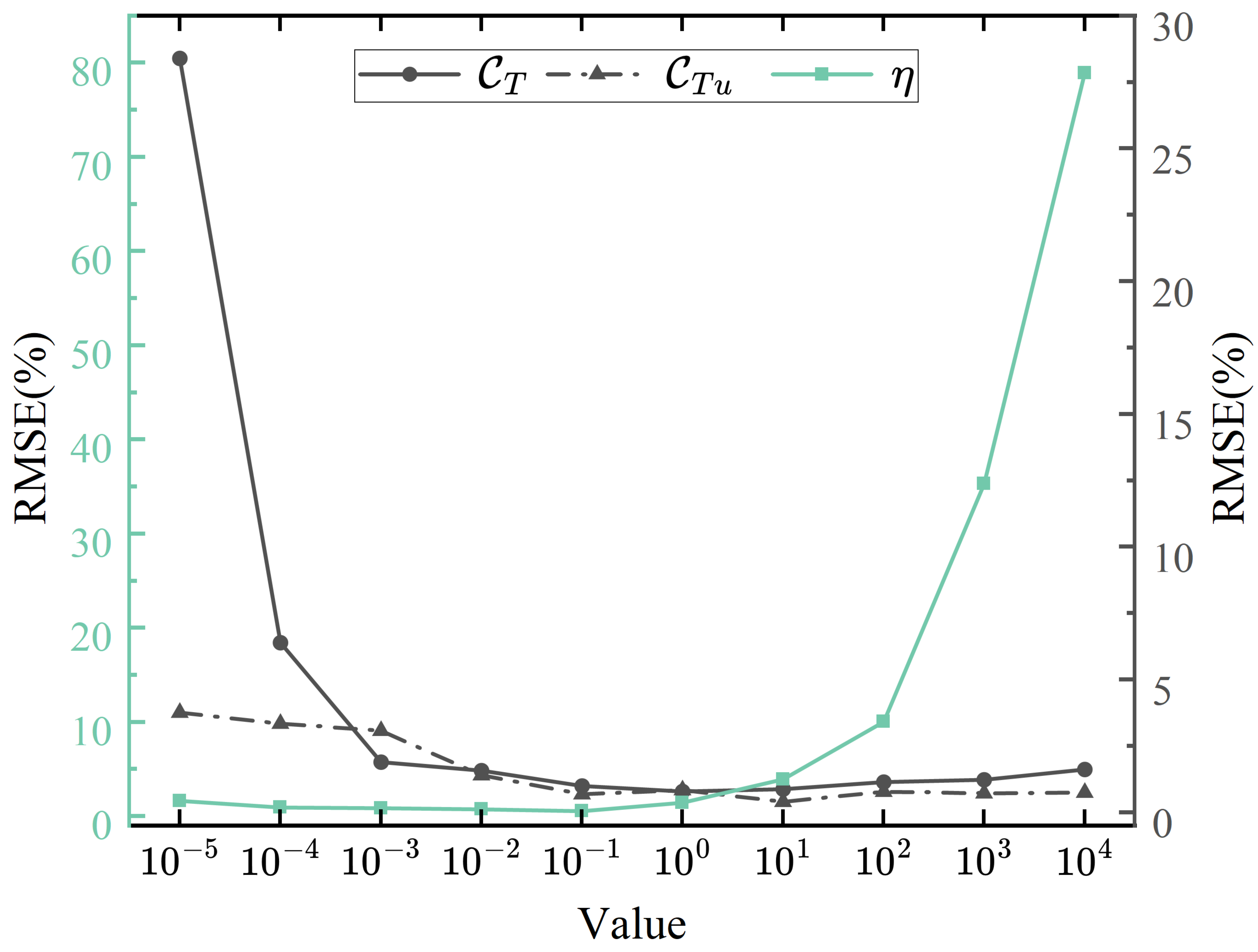}}
		
	}
	\caption{Parameter sensitivity analysis of CITL. (a) RMSE and (b) $L$ for different $L_{max}$ and $\varepsilon$. (c) RMSE for varying $\mathcal{C}_{T}$, $\mathcal{C}_{Tu}$, and $\eta$.}
	\label{Fig. 8}
	\vspace{-0.4cm}
\end{figure*}
In this part, the influence of node number on SOH estimation accuracy is investigated in MMD-LSTM-DA and CITL using the first 10 TL tasks shown in Table \ref{Table 3}. The hidden size of each layer of MMD-LSTM-DA is additionally set to 3, 6, 12, 24, 48, and 96, respectively. Undoubtedly, MMD-LSTM-DA requires retraining after every alteration in hidden size, after which the average results are documented. To obtain $RMSE$ of CITL after adding one node at a time, its experimental results of each task are selected based on the trial with the lowest error from 10 repeated trials. The experimental results are presented in Fig. \ref{Fig. 5}.

Fig \ref{Fig. 5}(a) illustrates the descent process of SOH error estimated by CITL as the node number increases to a maximum of 30. CITL results in the estimation error remaining below 1\% with a rapid damping across 7 TL tasks using no more than 10 nodes. Particularly, no more than 5 nodes are exclusively required for modeling the TL tasks of B5$\to$B1, B1$\to$B5, B5$\to$B4, and B5$\to$B2. For B3$\to$B1, approximately 20 nodes are utilized by CITL to keep SOH error under 1\%. As shown in Fig. \ref{Fig. 5}(b), the SOH error of MMD-LSTM-DA is indeed large when the total number of hidden nodes is initially set to 9. However, in each case, it is observed that an increase in the total number of hidden nodes leads to a decrease in estimation accuracy, such as from 9 to 18. In addition, its error can be less than 1\% utilizing a total of 18 nodes and less training time in B5$\to$B2. The overall trend of training time for MMD-LSTM-DA exhibits an upward trajectory as the number of nodes increases. Notably, Fig. \ref{Fig. 5}(b) simultaneously depicts that there is a decrease in training time with the addition of certain nodes in various cases, such as 18 nodes for B1$\to$B5, 18 nodes for B5$\to$B1, 36 nodes for B2$\to$B5, 18 nodes for B5$\to$B3, etc. These suggest that the impact of node number in deep TL-based models on estimation error lacks a clear regularity, resulting in uninterpretability for model structure in SOH estimation performance. Consequently, before developing a model for a particular SOH estimation task, it is challenging to achieve an optimal tradeoff between complexity and accuracy in deep learner models. Nevertheless, the estimation error of CITL does not suddenly rise after adding nodes. Meanwhile, CITL automatically adjusts the number of hidden nodes required for different TL tasks without artificial confusion for structure setting. Although the node number of CITL reaches $L_{max}$ in B2$\to$B5 and B5$\to$B3, its final estimation error approximates 1\%, with a significant accuracy improvement compared with MMD-LSTM-DA, as referred to Table \ref{Table 3}. Accordingly, the proposed approach can undertake the TL efficiency of each added node through CIL, thereby providing interpretability of the lightweight structure for TL-based battery SOH in AAVs.
\subsection{Impact of Labeled Sample Size}
Different degrees of labeled sample scarcity may be encountered in the real-world scenarios, potentially affecting the model performance of SOH monitoring. Thus, we conduct a comparative experiment by changing the number of labeled samples across $\{5, 10, 20, 30, 50\}$, in comparison with supervised/semi-supervised TL methods in target domain, including MMD-LSTM-DA, DDAN, BO-CNN-TL, and AS$^3$LSTM. The first TL task (B1$\to$B5) in Table \ref{Table 3} serves as the experimental case for this study. 

The average error of SOH estimation are illustrated in Fig. \ref{Fig. 88}. It suggests that the estimation accuracy generally improves with an increasing number of labeled samples, as these samples can supply more specific information on battery degradation. For BO-CNN-TL based on deep supervised TL, it demonstrates the poorest performance even when the sample size reaches its maximum. Note that training samples are labeled during the initial consecutive cycles rather than through random sampling over the entire life cycle. As the sample size increases from 20 to 50, the estimation accuracy of MMD-LSTM-TL and DDAN, both equipped with semi-supervised learning capabilities, reaches saturation. Analogously, the prediction error of CITL has not been significantly reduced between 30 and 50. However, when the labeled sample size is extremely small (e.g., 5, 10, 20, corresponding to about 0.5\%, 1\%, and 2\% of the total target samples, respectively), CITL consistently achieves superior monitoring performance compared to other methods.
\subsection{Parameter Sensitivity Analysis}
The hyper-parameters of CITL are mainly categorized into two categories. The first category encompasses maximum configurable number $L_{max}$ of hidden nodes and residual threshold $\varepsilon$, which directly govern the model structure and performance. The second category is composed of $\mathcal{C}_{T}$, $\mathcal{C}_{Tu}$, and $\eta$, all of which serve as regularization parameters for the objective function. Consequently, we take the B1$\to $B5 task as an illustrative example to investigate the influence of these parameters on the SOH prediction performance of CITL, prompting a sensitivity analysis.  

Initially, $\mathcal{C}_{T}$, $\mathcal{C}_{Tu}$, and $\eta$ are fixed at 1, 10, and 0.01, respectively. The $L_{max}$ and $\varepsilon$ are varied across $\{50,100,150,200,250\}$ and $\{10^{-1},10^{-2},10^{-3},10^{-4}\}$, respectively. The impact of $L_{max}$ and $\varepsilon$ on the RMSE and the actual number $L$ of used hidden nodes is explored, as depicted in Figs. \ref{Fig. 8}(a) and (b). As $\varepsilon$ is large and even though $L_{max}$ differs, CITL prematurely terminates the modeling with $L < {L_{\max }}$, and $L$ is approximately the same. As a result, the RMSE of SOH monitoring tends to be relatively high when $\varepsilon \in \{ 0.1,0.01\}$, as shown in Fig. 8(a). As $\varepsilon$  decreases, the SOH estimation accuracy of CITL is significantly improved. When $L_{max}=200$ and $\varepsilon=0.001$, this task achieves the lowest RMSE of approximately 0.6\%. Meanwhile, we can observe that the number of hidden nodes used reaches $L_{max}$ when $\varepsilon=10^{-4}$, since CITL requires more nodes for modeling to achieve a very small training error. 

Subsequently, the sensitivity analysis of $\mathcal{C}_{T}$, $\mathcal{C}_{Tu}$, and $\eta$ is conducted. $L_{max}$ and $\varepsilon$ are fixed at 200 and 0.001, respectively. The variable range of $\mathcal{C}_{T}$, $\mathcal{C}_{Tu}$, and $\eta$ are all $\{ {10^{-5}},{10^{-4}},{10^{-3}},{10^{-2}},{10^{- 1}},{10^0},{10^1},{10^2},{10^3},{10^4}\}$. Noteworthy, only one parameter among them is changing, while the other two parameters are fixed. When $\mathcal{C}_{T}$, $\mathcal{C}_{Tu}$, and $\eta$ need to be fixed, they are fixed at 1, 10, and 0.01, respectively. The RMSE curve of SOH estimation for varying $\mathcal{C}_{T}$, $\mathcal{C}_{Tu}$, and $\eta$ is illustrated in Fig. 8(c). $\mathcal{C}_{T}$ controls the significance of the labeled  data of target domain in the entire objective function. The RMSE is around 27\% when $\mathcal{C}_{T}$ is extremely small. As $\mathcal{C}_{T}$ increases to 1, the monitoring performance of CITL is improved rapidly. However, a excessively large $\mathcal{C}_{T}$ has no further positive impact on performance and may lead to overfitting. In contrast to $\mathcal{C}_{T}$, $\eta$ exhibits an opposite effect, under which CITL experiences a significantly large RMSE when $\eta = 10^{5}$. The optimal performance is obtained when $\eta = 0.1$. The influence of $\mathcal{C}_{Tu}$ on the SOH estimation accuracy also shows a similar trend, where a $\mathcal{C}_{Tu}$ that is too large or too small will both lead to an increase in RMSE. The optimal $\mathcal{C}_{Tu}$ is 10, with a RMSE of approximately 0.4\%.

\begin{figure*}[!t]
	\centering{
		\includegraphics[width=0.95\textwidth]{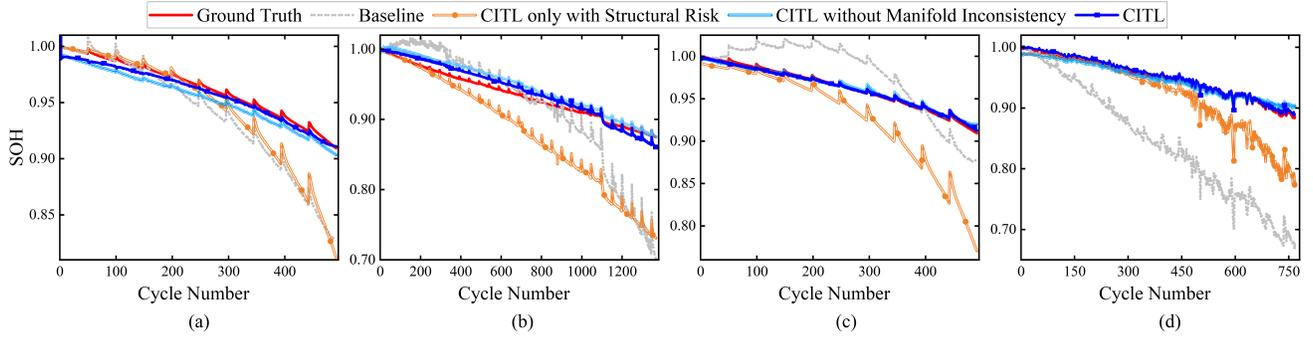}}
	\caption{SOH estimation results for ablation analysis in the TL task of (a) B1$\to$B9, (b) B9$\to$B1, (c) B8$\to$B9, and (d) B9$\to$B8.}
	\label{Fig. 7}
	\vspace{-0.4cm}
\end{figure*}
\subsection{Ablation Analysis}
\begin{table*}[!t]
	\centering
	\tabcolsep=8pt
	\small
	\caption{Ablation Analysis of the Proposed Approach}
	\resizebox{0.95\textwidth}{!}{ 
		\begin{tabular}{ccccccccc}
			\toprule
			\multirow{2}{*}{\textbf{TL Tasks}}&\multicolumn{2}{c}{\textbf{Baseline}}&\multicolumn{2}{c}{\textbf{CITL only with Structural Risk}}&\multicolumn{2}{c}{\textbf{CITL without Manifold Inconsistency}}&\multicolumn{2}{c}{\textbf{CITL}}\\
			\cmidrule(lr){2-3}\cmidrule(lr){4-5}\cmidrule(lr){6-7}\cmidrule(lr){8-9}
			&$R^2$&$RMSE$(\%)&$R^2$&$RMSE$(\%)&$R^2$&$RMSE$(\%)&$R^2$&$RMSE$(\%)\\
			\midrule
			B1$\to$B9&-&39.70$\pm$35.85&-&6.43$\pm$5.37&0.66$\pm$0.37&1.26$\pm$0.72&\textcolor{black}{\textbf{0.87$\pm$0.15}}&\textcolor{black}{\textbf{0.76$\pm$0.45}} \\
			B9$\to$B1&-&90.55$\pm$85.16&-&11.15$\pm$9.10&0.75$\pm$0.12&1.67$\pm$0.42&\textcolor{black}{\textbf{0.90$\pm$0.06}}&\textcolor{black}{\textbf{1.06$\pm$0.29}}\\
			B8$\to$B9&-&90.20$\pm$78.49&-&6.65$\pm$6.85&0.90$\pm$0.14&0.65$\pm$0.43&\textcolor{black}{\textbf{0.97$\pm$0.02}}& \textcolor{black}{\textbf{0.40$\pm$0.12}}\\
			B9$\to$B8&-&24.43$\pm$21.70&-&4.04$\pm$2.17&0.64$\pm$0.44&0.89$\pm$1.67&\textcolor{black}{\textbf{0.96$\pm$0.02}}&\textcolor{black}{\textbf{0.58$\pm$0.18}}\\
			\textbf{Average}&-&61.22$\pm$55.30&-&7.07$\pm$6.17&0.74$\pm$0.27&1.12$\pm$0.81&\textcolor{black}{\textbf{0.93$\pm$0.06}}&\textcolor{black}{\textbf{0.70$\pm$0.26}}\\
			\bottomrule
			\multicolumn{9}{l}{- indicates a failure in SOH prediction with ${R^2} < 0$}.
		\end{tabular}
	}
	\label{Table 4}
\end{table*}
An ablation analysis on CITL is performed to measure the impact of introduced terms on the performance for TL-based SOH estimation. The baseline and CITL only with structural risk represent the first and first two terms of \eqref{eq:9}, respectively. CITL without manifold inconsistency indicates the first two terms of \eqref{eq:8}. It is evident that CITL without manifold inconsistency and CITL possesses the capability of cross-domain learning, whereas others are exclusively trained using sparse data from the target battery. Meanwhile, AAVs are invariably subjected to drastic changes in real-world scenarios. Thus, we investigate two distinct categories of TL tasks: a) B1$\to$B9 and B9$\to$B1 represent large domain shifts under single-condition variation, and b) B8$\to$B9 and B9$\to$B8 denote substantial domain shifts involving multiple-condition variations.

Fig. \ref{Fig. 7} illustrates the SOH estimation curves, and their $R^2$ and $RMSE$ are reported in Table \ref{Table 4}. Baseline and structural risk are incapable of migrating aging knowledge, as evidenced by their consistent $R^2<0$ across entire tasks. Distinctly, the transferability of CITL under significant working condition shifts is demonstrated through an $R^2$ of 0.93 and average accuracy improvements of 98.86\% and 90.10\% compared to baseline and CITL without manifold inconsistency, respectively, with minimal standard deviation. Although CITL only with structural risk can only track the target trend in initial cycles, it is unable to incorporate data from other working conditions for subsequent accurate sensing. Thus, its SOH estimation gradually deviates from the ground truth as the cycle progresses. CITL without manifold term also demonstrates sound TL ability with accuracy gains of 86.90\%, 85.02\%, 90.23\%, and 77.97\% for B1$\to$B9, B9$\to$B1, B8$\to$B9, and B9$\to$B8, respectively, compared to CITL only with structural risk. Additionally, the effectiveness of the domain transfer constraint, represented by $\Gamma$, is confirmed by its close alignment with the aging trend of the target condition. Especially, it shows exceptional cross-domain learning capability in the large TL tasks between B8 and B9 conditions. In contrast to CITL without manifold inconsistency, the last term of CITL exhibits more stable distribution tracking and superior prediction performance, with a 25.68\% improvement in $R^2$ and 37.50\% reduction in $RMSE$. Therefore, the full use of unlabeled cycling data from a new condition improves estimation performance by exploring intrinsic spatial information.

\subsection{Model Evaluation on Embedded Hardware}
\begin{figure}[!t]
	\centering{
		\subfigure[]
		{\includegraphics[width=0.4\columnwidth]{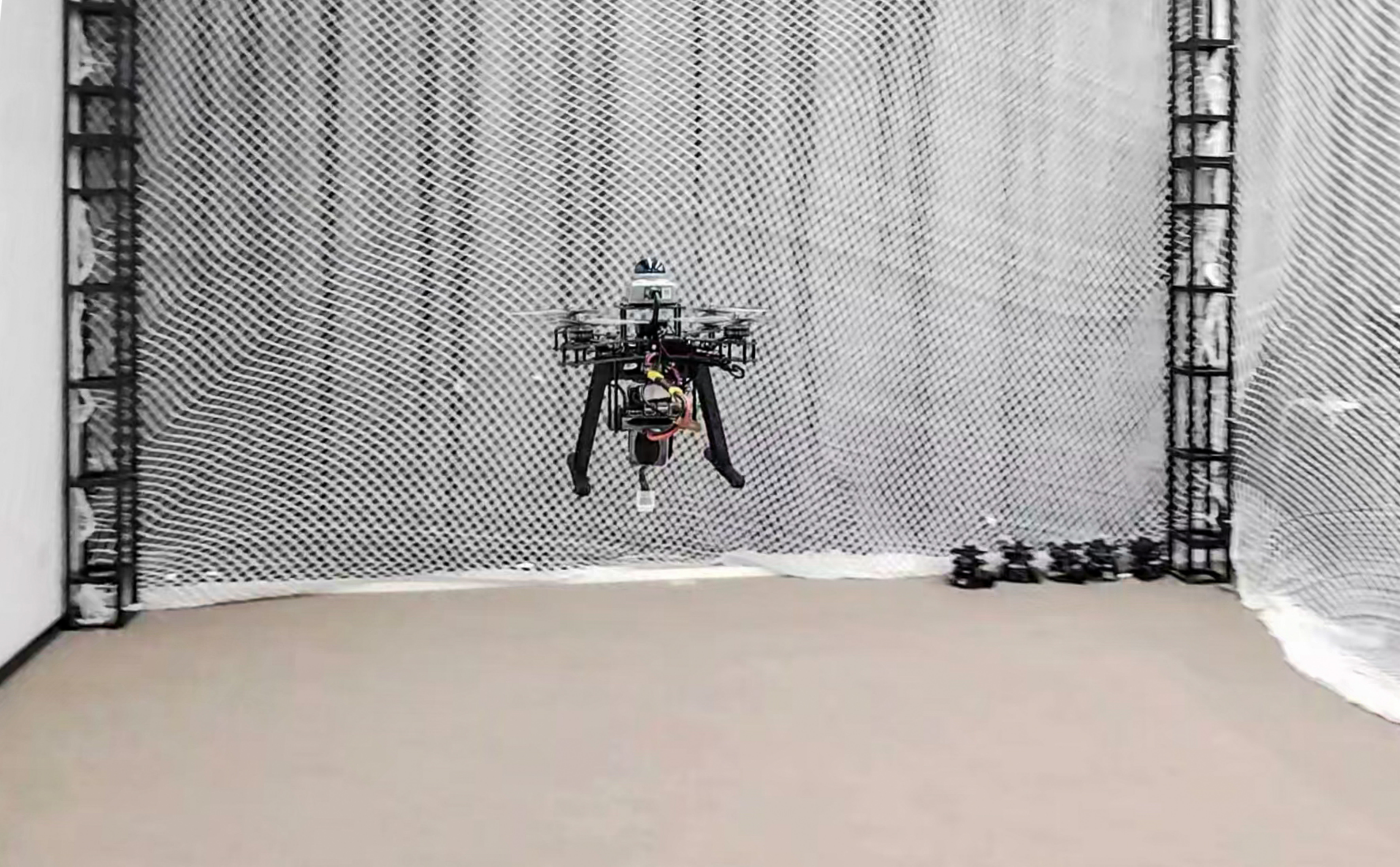}} \hfill
		\subfigure[]
		{\includegraphics[width=0.4\columnwidth]{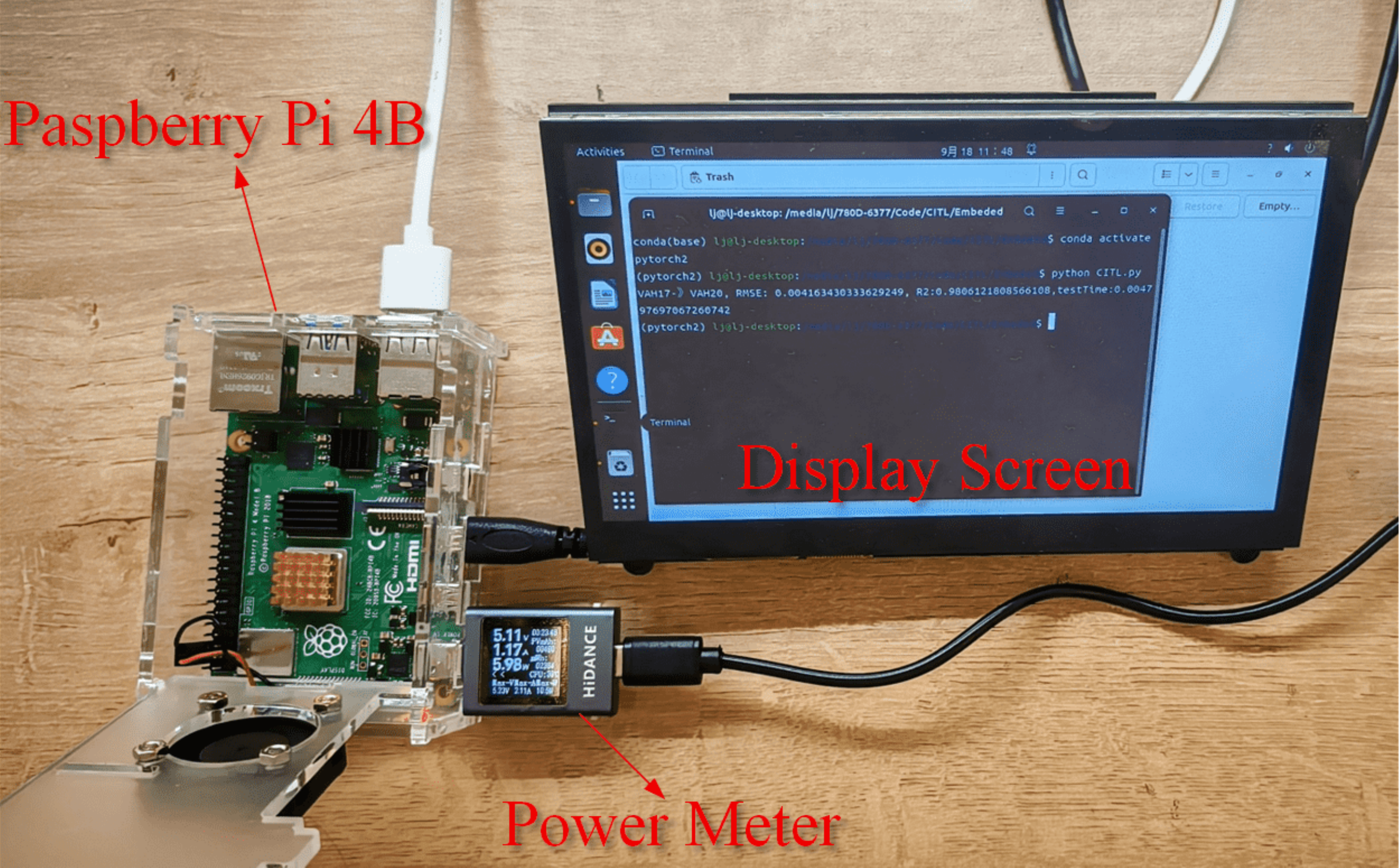}}
	}
	\caption{Developed embedded test platform with (a) Quadcopter and (b) Raspberry Pi 4B to verify CITL’s lightweight performance.}
	\label{Fig. 10}
	\vspace{-0.2cm}
\end{figure}
\begin{table}[!t]
	\centering
	\tabcolsep=2pt
	\caption{ Performance Comparisons of the B5$\to$B6 TL task on Embedded Hardware}
	\resizebox{0.9\columnwidth}{!}{ 
		\begin{tabular}{c|c|c|c|c|c c}
			\toprule
			\bf Model & \bf Parameter count & \bf Storage footprint & \bf Memory footprint &  \bf Inference time & \bf Power consumption &  \\
			\midrule
			SS-TCA \cite{li2020state}&5921&23.13 KB&17.01 MB&31.07 ms &1.47 W\\
			MMD-LSTM-DA \cite{han2022end}&50497&197.25 KB&33.20 MB&993.47 ms&3.03 W \\
			DDAN \cite{Ye2022StateofHealthEF}&263727&1030.18 KB&44.48 MB&10294.81 ms&2.75 W\\
			BO-CNN-TL \cite{zhang2023voltage}&101761&397.50 KB&31.13 MB&574.05 ms & 2.17 W\\
			AS$^3$LSTM \cite{JIANG2024Adaptive}&99298 &387.88 KB &35.57 MB&1579.80 ms&3.02 W \\
			CITL (Proposed)&936 &3.66 KB&13.47 MB&1.16 ms &1.15 W\\
			\bottomrule
	\end{tabular}}
	\label{tab:5}
\end{table}

To further evaluate the real-world applicability of CITL for AAV battery SOH monitoring, we comprehensively analyze its parameter count, storage footprint, memory footprint, inference time, and power consumption on a specific TL task (B5$\to$B6) using actual embedded hardware, in comparison with state-of-the-art approaches listed in Table \ref{Table 2}. The parameter count is obtained based on an input dimension of 102. The embedded test platform is illustrated in Fig. \ref{Fig. 10}, which features a research-oriented quadcopter equipped with a Raspberry Pi 4B. The Raspberry Pi serves as our standalone experimental platform for SOH monitoring, equipped with an ARM Cortex CPU (1.5 GHz, 4GB of RAM) running Ubuntu 22.04. All experiments are conducted using PyTorch 2 on this CPU. To fairly and accurately measure memory footprint and power consumption during the inference process, the average of their peak values over five repeated tests is used.

The performance comparisons on embedded hardware are presented in Table \ref{tab:5}. The size of the well-trained model is assessed by parameter count and storage footprint, with each parameter occupying 4 bytes. CITL occupies an extremely small storage space of 3.66 KB, as it is a fully connected network with a single hidden layer of about nine hidden nodes, the number of which is reported in Table \ref{Table 3}. However, SS-TCA requires additional storage for the reconstructed features. In the actual inference process, CITL also exhibits the lowest memory footprint, making it particularly suitable for embedded devices with limited memory and computational resources. Meanwhile, CITL achieves an inference time of 1.16 ms, which is 27×, 856×, 8875×, 495×, and 1362× faster than the five comparative models, respectively. The average peak power consumption of CITL is 1.15 W, which remains within acceptable ranges for most energy-constrained AAVs. In view of inference time and power consumption, it is evident that the computational cost of CITL is exceptionally low. Overall, CITL is highly suitable for onboard battery SOH monitoring in resource-constrained AAVs.
\section{Conclusion}
Considering the restricted computation resources in autonomous air vehicles (AAVs), a rapid and effective transfer learning (TL) approach is proposed for LiB state of health (SOH) monitoring for the target domain with limited labeled samples. The monitoring accuracy and speed of the proposed approach are improved by leveraging network node-based TL. Specifically, it is more efficient in applied contexts compared to other approaches through the following three achievements: 1) generating adequate hidden nodes with a high-quality cross-domain capability one by one for constructing the lightweight network; 2) residual feedback-based modeling of neural network using goal-optimized constraints on node parameters; 3) randomized feature extraction mechanism for obtaining end-to-end capability and exploring intrinsic geometrical information of unlabeled target data. With only 20 labeled cycles at the target condition, satisfying monitoring results are achieved by inheriting the consistent estimation ability from the source condition. The source estimator is also built using incremental transfer learning. The proposed approach has been exposed to extensive testing provided by a AAV battery dataset, and it has proven significant effectiveness as a powerful tool for SOH monitoring, particularly in scenarios with limited computing and energy resources of portable AAVs.

\bibliographystyle{IEEEtran}
\bibliography{references}
\appendices
\section{}\label{sec:theorem1}
According to (8) through (12), the optimization framework $\mathcal{J}({\boldsymbol{\beta} _{T,L}},{\bf{e}}_{Tl,L}^i,{\bf{e}}_{Tu,L}^j)$ of CITL is summarized as follows:
\begin{align}
	\mathcal{J}\!=&\frac{1}{2}\!\left\| {{{{\boldsymbol{\beta }}}'_{T,L}}} \right\|_2^2 \! +\! \frac{{{\mathcal{C}_T}}}{2}\!\sum\limits_{i = 0}^{{N_{Tl}}} {\left\| {{\mathbf{e}}_{Tl,L}^i} \right\|_2^2} \!+\! \frac{{{\mathcal{C}_{Tu}}}}{2}\!\sum\limits_{j = 0}^{{N_{Tu}}} {\left\| {{\mathbf{e}}_{Tu,L}^j} \right\|_2^2}  \nonumber\\
	&+\! \frac{\eta }{2}tr\left( {({{\bf{H}}_{T,L}^T{{{\boldsymbol{\beta}}}'_{T,L}}})^T{\mathbf{L}}{({{\bf{H}}_{T,L}^T{{{\boldsymbol{\beta}}}'_{T,L}}})}} \right).
	\label{eq:14}
\end{align}
where ${\mathbf{e}}_{Tu,L}^j = {\mathbf{H}}{_{S,Tu}^j}^T{{\mathbf{\beta }}_S} - {\boldsymbol{H}}{_{Tu,L}^j}^T{{\boldsymbol{\beta}'}_{T,L}}$. For ease of solving ${\boldsymbol{\beta }}_{T,L}$, $\mathcal{J}$ is rewritten as:
\begin{align}
	\!\!\!\!\mathcal{J} \!\!=&\frac{\left\| {{{\boldsymbol{\beta }}_T}} \right\|_F^2}{2} \!+\! \frac{\left\| {{{\boldsymbol{\beta }}_{T,L}}} \right\|_2^2}{2} \!\!+\!\! \frac{{{\mathcal{C}_T}\left\| {{{\bf{e}}_{Tl,L}}} \right\|_F^2}}{2} \!\!+ \!\!\frac{{{\mathcal{C}_{Tu}\left\| {{{\bf{e}}_{Tu,L}}} \right\|_F^2}}}{2} \nonumber\\
	&\!\!+ \!\!\frac{\eta }{2}\! tr \!\! \left( {{{\!({\bf{H}}_{T,L}^T \!{{\boldsymbol{\beta }}\!_T} \!\!+ \!{\bf{h}}_{T,L}^T{{\boldsymbol{\beta }}_{T,L}})}^T}\!\!{\bf{L}}({\bf{H}}_{T,L}^T \!{{\boldsymbol{\beta }}\!_T} \!\!+ \!{\bf{h}}_{T,L}^T{{\boldsymbol{\beta }}_{T,L}})\!} \right)\!.
\end{align}
${{{\boldsymbol{\beta }}_{T,L}}}$ can be solved by setting the derivative of $\mathcal{J}$ with respect to ${{{\boldsymbol{\beta }}_{T,L}}}$ to be zero by incorporating ${{\bf{e}}_{Tl,L}} = {{\bf{e}}_{Tl,L - 1}} - {{\bf{h}}^T_{Tl,L}}{{\boldsymbol{\beta }}_{T,L}}$ and ${{\bf{e}}_{Tu,L}} = {{\bf{e}}_{Tu,L - 1}} - {\bf{h}}_{Tu,L}^T{{\boldsymbol{\beta }}_{T,L}}$. Afterward, for $q = 1,\ldots,m$, we obtain
\begin{equation}
	{\beta _{T,L,q}} = \frac{{h{e_{T,L,q}}- \frac{\eta }{{{\mathcal{C}_T}}}{\left\langle {{\boldsymbol{\zeta} _{L - 1,q}},{{\bf{h}}_{T,L}}} \right\rangle }}}{{{\frac{1}{{{\mathcal{C}_T}}} + }{\left| h \right|_{T,L}} + \frac{\eta }{{{\mathcal{C}_T}}}{{\left\| {{{\bf{h}}_{T,L}}{{\bf{L}}^{1/2}}} \right\|}^2}}}
	\label{eq:23}
\end{equation}
where $\left\langle  \cdot  \right\rangle$ denotes the inner product of vectors, $h{e_{T,L,q}} = \left\langle {{{\bf{e}}_{Tl,L - 1,q}},{{\bf{h}}_{Tl,L}}} \right\rangle  + \frac{{{\mathcal{C}_{Tu}}}}{{{\mathcal{C}_T}}}\left\langle {{{\bf{e}}_{Tu,L - 1,q}},{{\bf{h}}_{Tu,L}}} \right\rangle $, ${{\boldsymbol{\zeta}}_{L - 1,q}} = {\bf{LH}}_{T,L}^T{{\boldsymbol{\beta }}_{T,q}}$, and ${\left| h \right|_{T,L}} = {\left\| {{{\bf{h}}_{Tl,L}}} \right\|^2} + \frac{{{C_{Tu}}}}{{{C_T}}}{\left\| {{{\bf{h}}_{Tu,L}}} \right\|^2}$. 

\textbf{Theorem 1}: Given three positive real numbers $({\mathcal{C}_T},{\mathcal{C}_{Tu}},\eta )$, $0 < r < 1$ and a non-negative real number sequence $\{ {\mu _L}\} $ with ${\lim _{L \to \infty }}{\mu _L} = 0$ and ${\mu _L} \leqslant 1 - r$, for $L = 1,2, \ldots $ and $q = 1, \ldots ,m$, the following definition is given as:
\begin{equation}
	{\delta _L} = \sum\limits_{q = 1}^m {{\delta _{L,q}}} ,{\text{ }}{\delta _{L,q}} = (1 - r - {\mu _L}){\left\| {{e_{Tl,L - 1,q}}} \right\|^2}.
\end{equation}
Denote $h'_{T,L} = \frac{1}{{{\mathcal{C}_T}}} + \frac{{{\mathcal{C}_{Tu}}}}{{{\mathcal{C}_T}}}{\left\| {{{\bf{h}}_{Tu,L}}} \right\|^2} + \frac{\eta }{{{\mathcal{C}_T}}}{\left\| {{{\bf{h}}_{T,L}}{{\bf{L}}^{1/2}}} \right\|^2}$ and $e{h_{Tl,L,q}} = \left\langle {{{\mathbf{e}}_{Tl,L - 1,q}},{{\mathbf{h}}_{Tl,L}}} \right\rangle $, if ${{\bf{h}}_{Tl,L}}$, ${{\bf{h}}_{Tu,L}}$, and ${{\bf{h}}_{T,L}}$ are generated to satisfy the following inequality: 
\begin{equation}
	{A_{L,q}} + {C_{L,q}} - {D_{L,q}} - {E_{L,q}} - {F_{L,q}} - {G_{L,q}} \geqslant {b^2_g}{\delta _{L,q}}
	\label{eq:25}
\end{equation}
where ${b_g} = h'_{T,L} + {\left\| {{{\mathbf{h}}_{Tl,L}}} \right\|^2}$
\begin{align}
	{A_{L,q}} =& \left( {\frac{1}{{{\mathcal{C}_T}}} + h'_{T,L} + {{\left\| {{{\mathbf{h}}_{Tl,L}}} \right\|}^2} + \frac{{{\mathcal{C}_{Tu}}}}{{{\mathcal{C}_T}}}{{\left\| {{{\mathbf{h}}_{Tu,L}}} \right\|}^2}} \right)eh_{_{Tl,L,q}}^2 \nonumber\\
	{C_{L,q}} =& \frac{{2{\mathcal{C}_{Tu}}}}{{{\mathcal{C}_T}}}h'_{T,L}e{h_{Tl,L,q}}\left\langle {{{\mathbf{e}}_{Tu,L - 1,q}},{{\mathbf{h}}_{Tu,L}}} \right\rangle \nonumber\\
	{D_{L,q}} = &\frac{{2\eta }}{{{\mathcal{C}_T}}}h'_{T,L}e{h_{Tl,L,q}}\left\langle {{{\boldsymbol{\zeta }}_{L - 1,q}},{{\mathbf{h}}_{T,L}}} \right\rangle \nonumber\\
	{E_{L,q}} =& \frac{{2\eta {\mathcal{C}_{Tu}}}}{{C_T^2}}h'_{T,L}\left\langle {{{\mathbf{e}}_{Tu,L - 1,q}},{{\mathbf{h}}_{Tu,L}}} \right\rangle \left\langle {{\boldsymbol{\zeta}_{L - 1,q}},{{\mathbf{h}}_{T,L}}} \right\rangle \nonumber\\
	{F_{L,q}} = &{\frac{\eta }{{{\mathcal{C}_T}}}^2}{\left\| {{{\mathbf{h}}_{Tl,L}}} \right\|^2}{\left\langle {{{\boldsymbol{\zeta }}_{L - 1,q}},{{\mathbf{h}}_{T,L}}} \right\rangle ^2} \nonumber\\
	{G_{L,q}} =& {\frac{{{\mathcal{C}_{Tu}}}}{{{\mathcal{C}_T}}}^2}{\left\| {{{\mathbf{h}}_{Tl,L}}} \right\|^2}{\left\langle {{{\mathbf{e}}_{Tu,L - 1,q}},{{\mathbf{h}}_{Tu,L}}} \right\rangle ^2}. \nonumber
\end{align}

Then, the output weight of $L^{th}$ hidden node are constructively evaluated by \eqref{eq:23}. Finally, ${\lim _{L \to \infty }}\left\| {{\mathcal{F}_{TL}} - {\mathcal{F}_{TL,L}}} \right\|=0$ is inferred. 

\textbf{Remark}: \textbf{Theorem 1} provides us with a constructive incremental scheme for shallow networks in term of TL, which brings a universal approximator in target domain. Eqs. \eqref{eq:23} and \eqref{eq:25} exactly represent our semi-supervised cross-domain learning mechanism that effectively leverages knowledge from both source and target domains to search for the appropriate input weight ${{\boldsymbol{\omega }}_{T,L}}$ and bias ${b_{T,L}}$ for a new hidden node in the TL process. To the best of our knowledge, the proposed learning mechanism is the first attempt in the territory of semi-supervised TL.
\section{} \label{sec:ap}
To theoretically guarantee the validity of transfer learning in CITL, this appendix yields a rigorous convergence analysis regarding \textbf{Theorem 1} and \textbf{Theorem 2}. Essentially, the goal of CITL is to construct a sound and lightweight SOH estimator toward the target battery with limited labeled data. Therefore, we can obtain the target residual $\left\| {{{\mathbf{e}}_{Tl,L}}} \right\|$ after adding each hidden node. In collaboration with the overall optimization framework, the convergence of $\left\| {{{\mathbf{e}}_{Tl,L}}} \right\|$ is primarily analyzed.

\textbf{Proof of Theorem 1}: Using (17), we have
\begin{align} \label{eq:30}
	&{\text{}}{\left\| {{{\mathbf{e}}_{Tl,L - 1}}} \right\|^2} - {\left\| {{{\mathbf{e}}_{Tl,L}}} \right\|^2} \nonumber\\
	= 	&\sum\limits_{q = 1}^m {\beta _{T,L,q}^2\left( 2{\eth_{T,L}}-{{{\left\| {{{\mathbf{h}}_{Tl,L}}} \right\|}^2}}\right)}  \nonumber\\
	&  + \! \sum\limits_{q = 1}^m {2{\beta _{T,L,q}}\!\left[ \!{\frac{\eta }{{{C_T}}}\left\langle {{{\mathbf{\varsigma }}_{L - 1,q}},{{\mathbf{h}}_{T,L}}} \right\rangle \! - \! \frac{{{C_{Tu}}}}{{{C_T}}}\left\langle {{{\mathbf{e}}_{Tu,L - 1,q}},{{\mathbf{h}}_{Tu,L}}} \right\rangle } \!\right]}  \nonumber\\
	= &\!\!\sum\limits_{q = 1}^m \!\! \left[\! A{'_{T,L}}{{\left\langle {{{\mathbf{e}}_{Tl,L - 1,q}},{{\mathbf{h}}_{Tl,L}}} \right\rangle }^2}
	\!-\! {{\left( {\frac{{\eta \left\| {{{\mathbf{h}}_{Tl,L}}} \right\|}}{{{C_T}{\eth_{T,L}}}}\left\langle {{\zeta _{L - 1,q}},{{\mathbf{h}}_{T,L}}} \right\rangle } \right)}^2} \right.& \nonumber\\
	& \left. + \frac{{2{\mathcal{C}_{Tu}}{\mathcal{A}_{T,L}}}}{{{C_T}}}\!\!\left\langle {{{\mathbf{e}}_{Tl,L - 1,q}},{{\mathbf{h}}_{Tl,L}}} \right\rangle\!\! \left\langle {{{\mathbf{e}}_{Tu,L - 1,q}},{{\mathbf{h}}_{Tu,L}}} \right\rangle  \right.& \nonumber\\
	& \left. - \frac{{2\eta {\mathcal{A}_{T,L}}}}{{{\mathcal{C}_T}}}\left\langle {{{\mathbf{e}}_{Tl,L - 1,q}},{{\mathbf{h}}_{Tl,L}}} \right\rangle \left\langle {{\zeta _{L - 1,q}},{{\mathbf{h}}_{T,L}}} \right\rangle  \right]  \nonumber\\
	& - \sum\limits_{q = 1}^m \left[ {{\left( {\frac{{{C_{Tu}}\left\| {{{\mathbf{h}}_{Tl,L}}} \right\|}}{{{C_T}{\eth_{T,L}}}}\left\langle {{{\mathbf{e}}_{Tu,L - 1,q}},{{\mathbf{h}}_{Tu,L}}} \right\rangle } \right)}^2}  \right.& \nonumber\\
	& \left. + 	\frac{{2\eta {\mathcal{C}_{Tu}}{\mathcal{A}_{T,L}}}}{{\mathcal{C}_T^2}}\left\langle {{{\mathbf{e}}_{Tu,L - 1,q}},{{\mathbf{h}}_{Tu,L}}} \right\rangle \left\langle {{\zeta _{L - 1,q}},{{\mathbf{h}}_{T,L}}} \right\rangle \right] \nonumber\\
	=& \!\sum\limits_{q = 1}^m \!{\frac{{{A_{L,q}} + {C_{L,q}} - {D_{L,q}} - {E_{L,q}} - {F_{L,q}} - {G_{L,q}}}}{{{b^2_{L,q}}}}  } \geqslant 0
\end{align}
where ${\mathcal{A}_{T,L}} = \left( {{\eth_{T,L}} - {{\left\| {{{\mathbf{h}}_{Tl,L}}} \right\|}^2}} \right)/\eth_{T,L}^2$, $\mathcal{A}'_{T,L} = {{\left( {2{\eth_{T,L}} - {{\left\| {{{\mathbf{h}}_{Tl,L}}} \right\|}^2}} \right)}}/{{\eth}^2_{T,L}}$, and ${\eth_{T,L}} = \frac{1}{{{\mathcal{C}_T}}} + {\left\| {{{\mathbf{h}}_{Tl,L}}} \right\|^2} + \frac{{{\mathcal{C}_{Tu}}}}{{{\mathcal{C}_T}}}{\left\| {{{\mathbf{h}}_{Tu,L}}} \right\|^2} + \frac{\eta }{{{\mathcal{C}_T}}}{\left\| {{{\mathbf{h}}_{T,L}}{{\mathbf{L}}^{1/2}}} \right\|^2}$. 

It is proved that the $\{ {\left\| {{{\mathbf{e}}_{Tl,L}}} \right\|^2}\} $ is monotonically decreasing. Then, we obtain
\begin{align}
	&{\left\| {{{\mathbf{e}}_{Tl,L}}} \right\|^2} - (\gamma  + {\mu _L}){\left\| {{{\mathbf{e}}_{Tl,L - 1}}} \right\|^2}  \nonumber\\
	=& \sum\limits_{q = 1}^m {\left[ {(1 - \gamma  - {\mu _L})\left\langle {{{\mathbf{e}}_{Tl,L - 1,q}},{{\mathbf{e}}_{Tl,L - 1,q}}} \right\rangle } \right]}  \nonumber \\
	&- \sum\limits_{q = 1}^m \left[ 2\left\langle {{{\mathbf{e}}_{Tl,L - 1,q}},{\mathbf{h}}_{T,L}^T{\beta _{T,L,q}}} \right\rangle  \right.\left. - \beta _{T,L,q}^2{\mathbf{h}}_{T,L}^{}{\mathbf{h}}_{T,L}^T  \right] \nonumber\\
	=&(1 - \gamma  - {\mu _L}){\left\| {{{\mathbf{e}}_{Tl,L - 1}}} \right\|^2} \nonumber\\
	&-\!\!\sum\limits_{q = 1}^m \!\!\left[\!\mathcal{A}'_{T,L}{{\left\langle {{{\mathbf{e}}_{Tl,L - 1,q}},{{\mathbf{h}}_{Tl,L}}} \right\rangle }^2} \!\!- \!{{\left( \!\!{\frac{{\eta \left\| {{{\mathbf{h}}_{Tl,L}}} \right\|}}{{{C_T}{\eth_{T,L}}}}\left\langle {{{\mathbf{\zeta }}_{L - 1,q}},{{\mathbf{h}}_{T,L}}} \right\rangle \!\!} \right)}^2} \right.& \nonumber\\
	& \left. + \frac{{2{C_{Tu}}{A_{T,L}}}}{{{C_T}}}\left\langle {{{\mathbf{e}}_{Tl,L - 1,q}},{{\mathbf{h}}_{Tl,L}}} \right\rangle \left\langle {{{\mathbf{e}}_{Tu,L - 1,q}},{{\mathbf{h}}_{Tu,L}}} \right\rangle \right.& \nonumber\\ 
	& \left. - \frac{{2\eta {A_{T,L}}}}{{{C_T}}}\left\langle {{{\mathbf{e}}_{Tl,L - 1,q}},{{\mathbf{h}}_{Tl,L}}} \right\rangle \left\langle {{{\mathbf{\zeta }}_{L - 1,q}},{{\mathbf{h}}_{T,L}}} \right\rangle \right] \nonumber\\ 
	&- \sum\limits_{q = 1}^m \left[\frac{{2\eta {C_{Tu}}{A_{T,L}}}}{{{C_T}{C_T}}}\left\langle {{{\mathbf{e}}_{Tu,L - 1,q}},{{\mathbf{h}}_{Tu,L}}} \right\rangle \left\langle {{\zeta _{L - 1,q}},{{\mathbf{h}}_{T,L}}} \right\rangle  \right.& \nonumber\\ 
	& \left. + {{\left( {\frac{{{C_{Tu}}\left\| {{{\mathbf{h}}_{Tl,L}}} \right\|}}{{{C_T}{_{T,L}}}}\left\langle {{{\mathbf{e}}_{Tu,L - 1,q}},{{\mathbf{h}}_{Tu,L}}} \right\rangle } \right)}^2} \right]\nonumber
\end{align}
\begin{align} \label{eq:31}
	=& {\delta _L} \!-\! \sum\limits_{q = 1}^m \!{\frac{{{A_{L,q}}\! + \!{C_{L,q}}\!-\! {D_{L,q}} \!- \!{E_{L,q}} \!- \!{F_{L,q}} \!- \!{G_{L,q}}}}{{{b^2_{L,q}}}}} \!\leqslant \! 0.
\end{align}

Finally, the following inequality is derived as:
\begin{align}
	0 \!\!\leqslant\!\! {\left\| {{{\mathbf{e}}_{Tl,L}}} \right\|^2} \!\!\leqslant \!\!(r \!+\! {\mu _L})\!{\left\| {{{\mathbf{e}}_{Tl,L}}} \right\|^2} \!\!\leqslant \!\! \prod\nolimits_{k = 1}^{L \!-\! 1} \!\!{(r \!+\! {\mu _k})\!{{\left\| {{{\mathbf{e}}_{Tl,0}}} \right\|}^2}} .
\end{align}
Then, ${\lim _{L \to \infty }}{\left\| {{{\mathbf{e}}_{Tl,L}}} \right\|^2} = 0$ is obtained because of ${\lim _{L \to \infty }}\prod\nolimits_{k = 1}^{L - 1} {(r + {\mu _k})} {\left\| {{{\mathbf{e}}_{Tl,0}}} \right\|^2}=0$, ${\lim _{L \to \infty }}{\left\| {{{\mathbf{e}}_{Tl,L}}} \right\|^2} = 0$, which implied ${\lim _{L \to \infty }}\left\| {{{\mathbf{e}}_{Tl,L}}} \right\| = 0$. This completes the \textbf{proof of Theorem 1}.

In the light of \textbf{Theorem 1} in Appendix \ref{sec:theorem1}, define the intermediate value $\tilde \beta _{T,L,q}$ as 
\begin{equation}
	{\tilde \beta _{T,L,q}} = \frac{{he_{T,L,q}^* - \frac{\eta }{{{\mathcal{C}_T}}}\left\langle {{\bf{\zeta }}_{_{L - 1,q}}^*,{{\bf{h}}_{T,L}}} \right\rangle }}{{\frac{1}{{{\mathcal{C}_T}}} + {{\left| h \right|}_{T,L}} + \frac{\eta }{{{\mathcal{C}_T}}}{{\left\| {{{\bf{L}}^{1/2}}{{\bf{h}}_{T,L}}} \right\|}^2}}}
\end{equation}
where ${\bf{\zeta }}_{L - 1,q}^*= {\bf{LH}}_{T,L}^T{\bf{\beta }}_{T,q}^*$, and $he_{T,L,q}^* = \left\langle {{\bf{e}}_{Tl,L - 1,q}^*,{{\bf{h}}_{Tl,L}}} \right\rangle  + \frac{{{\mathcal{C}_{Tu}}}}{{{\mathcal{C}_T}}}\left\langle {{\bf{e}}_{Tu,L - 1,q}^*,{{\bf{h}}_{Tu,L}}} \right\rangle $. Meanwhile, ${{\bf{\tilde e}}_{Tl,L}} = {\bf{e}}_{Tl,L - 1}^* - {\bf{h}}_{Tl,L}^T{\tilde {\boldsymbol{\beta }}_{T,L}}$, where ${{\mathbf{\tilde \beta }}_{T,L}} = [{\tilde \beta _{T,L,1}}, \ldots ,{\tilde \beta _{T,L,m}}]$.

\textbf{Proof of Theorem 2}: Due to ${\left\| {{\mathbf{e}}_{Tl,L}^*} \right\|^2} \leqslant {\left\| {{{{\mathbf{\tilde e}}}_{Tl,L}}} \right\|^2} = {\left\| {{\mathbf{e}}_{Tl,L - 1}^* - {\mathbf{h}}_{Tl,L}^T{{{\mathbf{\tilde \beta }}}_{T,L}}} \right\|^2} \leqslant {\left\| {{{{\mathbf{\tilde e}}}_{Tl,L - 1}}} \right\|^2}$, we can have 
\begin{align} \label{eq:33}
	&{\left\| {{\mathbf{e}}_{Tl,L}^*} \right\|^2} - {\left\| {{\mathbf{e}}_{Tl,L - 1}^*} \right\|^2} \leqslant {\left\| {{{{\mathbf{\tilde e}}}_{Tl,L}}} \right\|^2} - {\left\| {{\mathbf{e}}_{Tl,L - 1}^*} \right\|^2} \nonumber\\
	=& \!\sum\limits_{q = 1}^m \!{\left[ {\tilde \beta _{T,L,q}^2{\mathbf{h}}_{Tl,L}{\mathbf{h}}^T_{Tl,L}} \right.} \left. \!{\!-2{{\tilde \beta }_{T,L,q}}\left\langle {{\mathbf{e}}_{Tl,L - 1,q}^*,{\mathbf{h}}_{Tl,L}^T} \right\rangle } \right]. 
\end{align}

Similar to \eqref{eq:30}, \eqref{eq:33} is rewritten as:
\begin{align}\label{eq:34}
	&{\text{ }}{\left\| {{\mathbf{e}}_{Tl,L - 1}^*} \right\|^2} - {\left\| {{\mathbf{e}}_{Tl,L}^*} \right\|^2}  \nonumber\\ 
	\geqslant& \sum\limits_{q = 1}^m \left[ 2\left\langle  {{\mathbf{e}}_{Tl,L - 1,q}^*,{\mathbf{h}}_{Tl,L}^T{{\tilde \beta }_{T,L,q}}} \right\rangle \right. \left.  - {\tilde \beta _{T,L,q}^2{{\mathbf{h}}_{Tl,L}}{\mathbf{h}}_{Tl,L}^T} \right]  \nonumber\\ 
	=& \! \sum\limits_{q = 1}^m \!\frac{{A_{L,q}^* + C_{L,q}^* - D_{L,q}^* - E_{L,q}^* - F_{L,q}^* - G_{L,q}^*}}{{{b^2_{L,q}}}} \geqslant 0.
\end{align}
Subsequently, referred to \eqref{eq:31}, we can infer
\begin{align}\label{eq:35}
	&{\left\| {{\mathbf{e}}_{Tl,L}^*} \right\|^2} - (r + {\mu _L}){\left\| {{\mathbf{e}}_{Tl,L - 1}^*} \right\|^2} \nonumber \\
	\leqslant &{\left\| {{{{\mathbf{\tilde e}}}_{Tl,L}}} \right\|^2} - (r + {\mu _L}){\left\| {{\mathbf{e}}_{Tl,L - 1}^*} \right\|^2} \nonumber\\ 
	= &(1 - r - {\mu _L}){\left\| {{\mathbf{e}}_{Tl,L - 1,q}^*} \right\|^2} \nonumber\\ 
	&- \sum\limits_{q = 1}^m \left[2\left\langle {{\mathbf{e}}_{Tl,L - 1,q}^*,{\mathbf{h}}_{Tl,L}^T{{\tilde \beta }_{T,L,q}}} \right\rangle  \right.\left. - {\tilde \beta _{T,L,q}^2{{\mathbf{h}}_{Tl,L}}{\mathbf{h}}_{Tl,L}^T} \right] \nonumber\\
	= &\delta _L^* \!-\! \sum\limits_{q = 1}^m \!{\frac{{A_{L,q}^* \!+\! C_{L,q}^* \!-\! D_{L,q}^* \!-\! E_{L,q}^* \!-\! F_{L,q}^* \!-\! G_{L,q}^*}}{{{b^2_{L,q}}}}}\! \leqslant \!0.
\end{align}

According to \eqref{eq:34}, $ {\left\| {{\mathbf{e}}_{Tl,L - 1}^*} \right\|} $ is monotonically decreasing with $L \to \infty $. And referring to \eqref{eq:35}, we can obtain
\begin{align}
	0 \!\!\leqslant \!\!{\left\| {{{\mathbf{e}}^*_{Tl,L}}} \right\|^2} \!\!\leqslant\!\! (r \!+\! {\mu _L})\!{\left\| {{{\mathbf{e}}^*_{Tl,L}}} \right\|^2}\!\!\!\leqslant\!\! \prod\nolimits_{k = 1}^{L \!-\! 1}\!\!\!{(r \!+\! {\mu _k})\!{{\left\| {{\mathbf{e}}_{Tl,0}^*} \right\|}^2}}.
\end{align}
Similarly, the \textbf{proof of Theorem 2} is completed with ${\lim _{L \to  + \infty }\left\| {{\mathbf{e}}_{Tl,L}^*} \right\| } = 0$.

\textbf{Remark}: The aforementioned proofs provide constructive incremental semi-supervised TL for target domain with a strong theoretical guarantee. It confirms that the prediction residual of target data gradually decreases as the number of nodes increases, and domain transfer is positively effective in an explainable fashion compared with deep learning-based TL. Hidden nodes capable of facilitating effective cross-domain knowledge acquisition are subsequently generated with a high degree of compactness. As a result, a lightweight TL approach is produced for our battery SOH estimation in portable mobile devices. Generally, \textbf{Theorem 2}, an improved variant of \textbf{Theorem 1}, is used as the acquiescent algorithmic foundation for CITL. 
\end{document}